\documentclass{article}

\usepackage{graphicx}
\usepackage{cite}
\usepackage{amsmath}

\newcommand{\widefigwidth}{5.0in}
\newcommand{\figwidth}{3.5in}
\newcommand{\mfigwidth}{3.0in}

\usepackage{xcolor} 

\newcommand{\eq}[1]{Eq.~(\ref{eq.#1})} 
\newcommand{\fig}[1]{Fig.~\ref{fig.#1}}

\newcommand{\tbl}[1]{Table~\ref{table.#1}}

\newcommand{\sect}[1]{Section~\ref{sect.#1}}
\newcommand{\sectA}[1]{Appendix~\ref{sect.#1}}
\newcommand{\sectlabel}[1]{\label{sect.#1}}
\newcommand{\eqlabel}[1]{\label{eq.#1}}
\newcommand{\figlabel}[1]{\label{fig.#1}}
\newcommand{\tbllabel}[1]{\label{table.#1}}

\newcommand{\Oxygen}{\ensuremath{\mbox{O}_2}}
\newcommand{\Doxygen}{D_{\textnormal{\Oxygen}}} 

\newcommand{\rVessel}{\ensuremath{r}} 
\newcommand{\vVessel}{\ensuremath{v}}  
\newcommand{\lVessel}{\ensuremath{\ell}} 
\newcommand{\tVessel}{\ensuremath{t}} 
\newcommand{\VVessel}{\ensuremath{V}} 
\newcommand{\FVessel}{\ensuremath{F}} 
\newcommand{\RVessel}{\ensuremath{R}} 

\newcommand{\viscosity}{\ensuremath{\eta}}
\newcommand{\bloodVolume}{\ensuremath{V_{\textnormal{blood}}}}
\newcommand{\bloodFlow}{\ensuremath{F_{\textnormal{blood}}}}
\newcommand{\bloodAvgTransit}{\ensuremath{t_{\textnormal{blood}}}} 

\newcommand{\rRobot}{\ensuremath{r_{\textnormal{robot}}}} 

\newcommand\Pec{\mbox{Pe}}  

\newcommand{\meter}{\mbox{m}}
\newcommand{\millimeter}{\mbox{mm}}
\newcommand{\micron}{\mbox{$\mu$m}}
\newcommand{\nanometer}{\mbox{nm}}
\newcommand{\liter}{\mbox{L}}
\newcommand{\milliliter}{\mbox{mL}}
\newcommand{\minute}{\mbox{min}}
\newcommand{\second}{\mbox{s}}

\newcommand{\mmHg}{\mbox{mmHg}} 
\newcommand{\vresistance}{\mbox{MPa s}/\mbox{m}^3}

\newcommand{\kilowatt}{\mbox{kW}}

\newcommand{\picowatt}{\mbox{pW}}

\newcommand{\molecule}{\mbox{molecule}}

\title{Chemical Power Variability among Microscopic Robots in Blood Vessels}
\author{Tad Hogg\\
{\small Institute for Molecular Manufacturing}\\{\small Palo Alto, CA}
}

\begin{document}
\maketitle

\begin{abstract}

Fuel cells using oxygen and glucose could power microscopic robots operating in blood vessels. Swarms of such robots can significantly reduce oxygen concentration, depending on the time between successive transits of the lung, hematocrit variation in vessels and tissue oxygen consumption. These factors differ among circulation paths through the body. This paper evaluates how these variations affect the minimum oxygen concentration due to robot consumption and where it occurs: mainly in moderate-sized veins toward the end of long paths prior to their merging with veins from shorter paths. This shows that tens of billions of robots can obtain hundreds of picowatts throughout the body with minor reduction in total oxygen. However, a trillion robots significantly deplete oxygen in some parts of the body. By storing oxygen or limiting their consumption in long circulation paths, robots can actively mitigate this depletion. The variation in behavior is illustrated in three cases: the portal system which involves passage through two capillary networks, the spleen whose slits significantly slow some of the flow, and large tissue consumption in coronary circulation. 

\end{abstract}

\section{Introduction}

Small implanted medical devices can provide high precision treatments~\cite{dong07,nelson10,rajendran24}.
However, there are numerous challenges to realizing their potential to improve medicine,
including providing power~\cite{amar15,bazaka13}.

Power is a particular challenge for microscopic robots~\cite{freitas99,morris01} operating in blood vessels due to their small sizes, large numbers, distribution throughout the body and continual movement. 
Chemical power is one option that is especially useful for long-term operations where it may not be convenient or feasible to deliver power to patients from outside the body or with implanted batteries. An example is when robots monitor for rare conditions and must be able to respond when and where those conditions occur. 
 
Chemical power involves reacting oxygen with a fuel source.
Important questions are the amount of power and how this oxygen consumption affects oxygen available for tissue. 
An important aspect of addressing these questions is the considerable variation in circulation paths~\cite{morris57}. Accounting for such variation is necessary to determine the range of behaviors~\cite{galton1883}, especially extreme cases~\cite{savage09}. 
In particular, this variation affects the minimal oxygen concentration and where it occurs, which are important for accessing the safety of robots using chemical power~\cite{carreau11}.

An important aspect of evaluating the effect of oxygen-consuming robots is the considerable variation in circulation paths~\cite{morris57}. Accounting for such variation in important for determining the range of behaviors, especially extreme cases~\cite{savage09}. 
An important criterion for robot operation and the effect on tissue is the minimum oxygen concentration anywhere in the body due to robot consumption. The minimum value is important both to avoid tissue hypoxia and to determine a lower bound on the power available to robots.
Determining the minimal oxygen due to robot consumption and where it occurs are important to assess the safety of robot power use in different tissues~\cite{carreau11}.

After describing studies of power for small devices in \sect{related}, this paper addresses these questions with a model of how robots consume oxygen while circulating through different parts of the body.
Specifically,  \sect{variation} motivates the model developed here by discussing the major variations in circulation loops and how they affect oxygen concentration.  \sect{variation model} describes a network circulation model accounting for these variations. \sect{parameters} provides estimates for the parameters of each segment of this network. The model's predictions of robot power and oxygen depletion in different parts of the body is discussed in \sect{robot power}. Additional details of some circulation loops that differ significantly from the average circulation are given in \sect{some circulation paths}. Based on this model, \sect{mitigation} provides mitigation strategies robots could use to reduce their effect on oxygen levels arising from variation in circulation loops. \sect{discussion} summarizes the results from this study. Finally, \sect{conclusion} discusses limitations and extensions to the model.

\section{Related Work}\sectlabel{related}

The design and behavior of tiny medical devices has received considerable attention.
These include devices powered and controlled from outside the body as well as autonomous robots that harvest energy locally and use on-board logic to determine their actions~\cite{brooks20}.
External power and control with magnetic fields has been demonstrated for a variety of devices~\cite{dreyfus05,martel07}.
These fields can apply forces to the devices to move them in specific directions under external control.

For autonomous robots, chemical power derived from fuel cells can provide power from their environments without requiring external energy sources.
One example is millimeter-size fuel cells that use lactate as the fuel and enzymes to catalyze reactions. These can achieve high efficiency for microwatt power~\cite{hashemi20}. 
For mobile microscopic devices in blood vessels, glucose-oxygen fuel cells~\cite{an11,chaudhuri03,gogova10,rapoport12,zebda13} are an appealing power source due to widespread and ongoing availability of these chemicals in the blood.

In terms of how robot oxygen consumption affects oxygen concentration, a major constraint is that a robot cannot collect oxygen faster than it diffuses to its surface. This means that a few tiny robots have little effect on overall oxygen concentration in the blood even if they consume all the oxygen reaching them~\cite{hogg10}. However, a large number of robots could consume a significant portion of the oxygen. 
A previous study~\cite{hogg21-average} evaluated how large numbers of robots affect the oxygen concentration by modeling an average circulation loop where blood takes about a minute to travel from the lungs, through the body and back to the lungs. This estimated available chemical power as a function of the number of robots in the blood and found that the minimal oxygen concentration occurs at the end of the loop, just before blood reaches a lung capillary.  However, by assuming an average circulation loop, this model did not account for variation in the circulation.

\section{Major Variations Affecting Oxygen Consumption}\sectlabel{variation}

In the blood, oxygen concentration is much lower than that of glucose~\cite{hogg10}. Thus, we focus on oxygen to evaluate the amount of power available to robots from glucose-oxygen fuel cells.

When microscopic robots use oxygen for power in the bloodstream, oxygen concentration in blood plasma changes due to consumption by robots and tissue, and replenishment from red cells. Thus the key physiological properties that determine how robots affect oxygen concentration in a vessel are the time blood takes to travel through the vessel, the fraction of the volume occupied by cells, i.e., the hematocrit, and tissue power demand. 
Evaluating these factors for a typical circulation loop, illustrated in \fig{average circulation}, determines the typical power available to robots and their systemic effect on oxygen available to tissue~\cite{hogg21-average}.
However, determining how these effects vary with location in the body requires accounting for major differences in circulation paths.
To address this variation, this section describes the major relevant variations in transit time, hematocrit and tissue demand. 

\begin{figure}[t]
\centering 
\includegraphics[width=\widefigwidth]{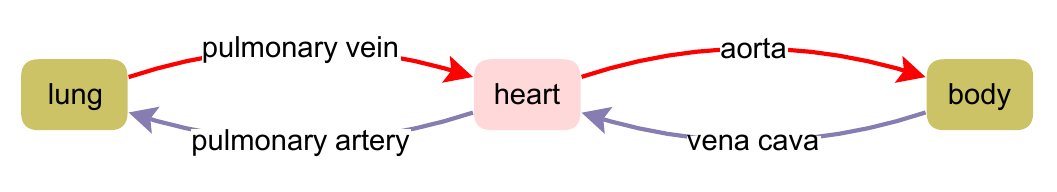}
\caption{Circulation from lungs to the rest of the body and back to the lungs.}\figlabel{average circulation}
\end{figure}

\subsection{Transit Time Variation}\sectlabel{transit variation}

Blood flow responds to changing activity during a day, e.g., resting, exercise or eating. Over seconds, flow varies as the heart beats. The evaluation of robot power considered here is over the course of a medical treatment, i.e., minutes to hours for a resting patient. This time scale allows averaging over the heart cycle and using a resting tissue demand.
Under these conditions, the main variation from circulation arises from differences in the circulation time to various parts of the body~\cite{morris57}. 

Shorter transit times, e.g., through tissue close to the heart, have less time for robots to deplete oxygen in the blood. Conversely, longer transit times, e,g., to the lower extremities, have more oxygen depletion than average, until the returning blood merges with flows from shorter circuits on its way back to the heart.
These circulation loops differ by tens of centimeters in distance and tens of seconds in transit time. These variations are comparable to the average circulation, e.g., the one-minute average circulation time, which corresponds to the time the heart takes to pump a volume equal to the entire blood volume. 
Thus these differences are significant fractions of the average circulation and consequently can give large differences in robot oxygen consumption.

Additional sources of variation arise over smaller distances. For instance, connected arteries and veins (anastomosis) give multiple flow paths to the same location in the body. If these paths have significantly different transit times, they could mix blood with different oxygen concentrations due to robots consuming oxygen for different amounts of time in the arteries and veins, whereas there would not be any difference along these paths due to oxygen consumption by tissue, which only occurs in capillaries. However, unless one of the paths is nearly occluded, the difference in transit time between these connected paths is considerably less than the variation due to flow to different parts of the body. 

Flow though small vessels can have significant variation in transit time even within a single part of the circulation.
For instance, measurements of flow through resting leg muscle give a range of transit time from about 2 to 4 minutes~\cite{mizuno03}.
This is due to measuring transit of blood, which includes a portion of the blood plasma leaving capillaries at their high pressure arterial side and returning at the low pressure venous side~\cite{feher17}. 
This pooling in interstitial fluid affects the transit time of plasma, but cells and robots are confined to the vessels, so their transit time does not include a contribution from the slow motion of some of the blood fluid that passes out of and back into a capillary.
This is an example of the difficulty in relating reported perfusion measurements to how robots move with the blood flow.
Additional interpretation challenges arise from perfusion measurements that are reported per unit mass of tissue: whether that includes the mass of blood in the tissue or some surrounding structures~\cite{peters18}.

Additional small-scale variation arises from changing flow in capillary networks.
For example, collapsed capillaries in the upper part of the lungs when standing~\cite{feher17} changes the flow compared to a prone position.
Even when the body is in a fixed position, flow through capillaries can vary with time as  pre-capillary sphincters open and close, so that, at rest in some tissues, only about $10$ to $20\%$ of capillaries are open at a time~\cite{krogh22,feher17}.
Robots stuck at pre-capillary sphincters may need to wait for them to open, during which time the robots can continue to consume oxygen. A similar effect occurs when flow through a capillary is temporarily slowed as a large white blood cell passes through~\cite{freitas99}.
The extent to which this variation in microflows matters for oxygen consumption depends on how often and how long such blockages occur, in particular the extent to which this variation results in a more gradual variation in capillary flow than could be the case if most capillaries in resting tissue are completely closed~\cite{angleys20,poole13}.

For robot oxygen use, differences in distances to merging vessels of more than a few centimeters could significantly affect the level and location of minimum oxygen concentration in long loops that merge with shorter ones.
Accounting for these various circulation times is necessary to determine the minimal oxygen concentration in the circulation due to robot consumption and where it occurs.
Variations over smaller distances are less significant. This is because any microscale variation in transit time, even if large on a percentage basis, will be small in absolute value, so there is little difference in the amount of oxygen consumed by robots among those variations. Thus the model described here focuses on large-scale differences in transit times as most relevant to robot oxygen consumption.

\subsection{Hematocrit Variation}\sectlabel{hematocrit variation}

Red blood cells replenish oxygen removed from the plasma by robots or tissue. The fraction of blood volume occupied by cells, the hematocrit, varies with vessel size~\cite{freitas99}.  
Typically, small vessels have lower hematocrit, which is known as the Fahraeus effect~\cite{mcHedlishvili87}. This arises from cells moving a bit faster than plasma in these vessels.
The reduced hematocrit in small vessels affects how much robots alter oxygen concentration in two competing ways: lower hematocrit means there are fewer cells to replenish oxygen removed from the plasma while faster speed relative to the plasma increases the rate at which new, less depleted cells reach the capillary. 

Hematocrit is typically around $45\%$ in large vessels, $33\%$ in capillaries and has intermediate values between these sizes~\cite{freitas99}. 
However, there is considerable variation among small vessels in different organs~\cite{gibson46,linderkamp80}.
An extreme example of hematocrit variation is in parts of the spleen, where cells move slower than plasma (i.e., an inverse Fahraeus effect), giving hematocrit of up to 80\%~\cite{wadenvik88,macdonald91}. This slow motion is due to the spleen filtering the blood by passing blood cells through small slits.

Blood usually passes through a single capillary network during its circulation through the body. 
Portal flows are an exception in which blood passes through two separate capillary networks before returning to the heart. Robots in such flows encounter low hematocrit in each capillary network, separated by transit through larger vessels with larger hematocrit. Tissue consumes oxygen during both capillary transits. 
The main instance of portal flow is through digestive organs and the liver. This accounts for a significant fraction of the total blood flow so is an important source of variation in hematocrit experienced by the robots.
Other portal systems are in the kidneys  
and the hypophyseal portal system in the brain~\cite{feher17}.
These portal systems involve much smaller flows over shorter distances than blood flowing through the liver so are not considered separately in the model discussed here.

\subsection{Tissue Power Demand Variation}\sectlabel{tissue variation}

Organs and cells within those organs vary in their oxygen requirements, even at their basal level of activity~\cite{freitas99}. 
Tissue extracts oxygen from blood as it passes through capillaries. Thus the relevant variation is that of tissue power demand \emph{per capillary} rather than tissue power density (i.e., power per unit volume). The consumption from a capillary is the product of the volume of cells supplied by that capillary and the tissue power density in that volume. These two factors tend to vary in opposite directions because
variation in basal power demand correlates with variation in capillary density in tissue~\cite{freitas99}.
Thus tissue with higher power density tends to have fewer cells supplied by each capillary, so that
tissue oxygen consumption per capillary does not vary as much as power density. 

Moreover, unless robots significantly reduce or avoid consumption in large vessels, most of the robot oxygen consumption occurs while they are in large vessels, rather than passing through a capillary. Thus the oxygen reduction by tissue and robots during a capillary transit is a relatively minor part of the overall reduction due to robots as they travel around a typical circulation loop. E.g., even increasing tissue consumption by a factor of 2 would not significantly alter the model results.

These considerations suggest variation in resting power demand among most tissues in the body will not significantly alter the consequences of robot oxygen consumption.
There are two exceptional cases where variation in tissue consumption could be more significant.
The first exception is the coronary circulation. Heart muscle consumes about $70\%$ of the oxygen in the blood passing through it, even at resting demand~\cite{deussen12}. Moreover, this is a relatively short circulation loop so the one-second capillary transit accounts for a significant fraction of the total circuit transit time and is associated with higher than average tissue demand.
The second exception is when blood spends much more than one second in capillary networks. This occurs in the slowly moving blood within the spleen described in \sect{hematocrit variation}. The spleen filters more blood than required to support its metabolic requirements, corresponding to lower power demand per capillary.

\section{Modeling Variation in Blood Flow}\sectlabel{variation model}

Accounting for the variations discussed in \sect{variation} requires separating the circulation into a network of segments, each of which models an aggregation of paths through the circulation with similar transit times, hematocrit and tissue demand. Such a network extends the average circulation loop shown in \fig{average circulation} to account for the major variations relevant to robot oxygen consumption. This includes variations in distance, time and speed of flow, and how blood returning from these locations mixes in larger veins before returning to the heart. The model accounts for how flow splits and merges among these segments, and how those processes affect oxygen concentration in the blood. 

The network model consists of segments that aggregate circulation paths with similar transit times. The flow properties of the vessels aggregated into a segment combine to give those properties for the segment as a whole. For example, the blood volume and flow rate of a segment is the sum of volumes and flows of the vessels in that segment.
A segment can then be modeled as having a single averaged transit time and hematocrit profile that is representative of the vessels aggregated into that segment.

The remainder of this section describes the processes determining flow and concentration at vessel junctions and the choice of segment network to evaluate the variation in oxygen consumption.

\subsection{Blood Flow Properties}

Pressure differences drive the flow of blood through vessels. Each vessel has an associated pressure drop $\Delta P$ (averaged over the heart cycle), vascular resistance \RVessel, flow rate \FVessel, blood volume \VVessel\ and average transit time \tVessel. 
In analogy with Ohm's law for electrical circuits, vascular resistance relates pressure drop to flow in the vessel. In addition, the average transit time relates the flow rate and volume of blood in a vessel. These relations are
\begin{equation}\eqlabel{vessel}
\begin{split}
\Delta P &= \RVessel \FVessel \\
\VVessel &= \tVessel \FVessel
\end{split}
\end{equation}

These quantities depend on the geometry of the vessels. One example is a straight vessel of radius \rVessel\ and length \lVessel. In this case, average speed in the vessel is $\vVessel = \FVessel/(\pi \rVessel^2)$ and transit time is $\tVessel = \lVessel/\vVessel$.
Poiseuille's relation~\cite{happel83} between laminar flow and pressure drop in a straight vessel gives vascular resistance
\begin{equation}
R = \frac{8 \lVessel \viscosity}{\pi \rVessel^4}
\end{equation}
where \viscosity\ is the blood's viscosity. 
Due to the fourth power of radius in the denominator, resistance increases rapidly as vessel size decreases. The organization of the circulatory system is a trade-off between using large vessels to reduce vascular resistance and small vessels to efficiently exchange chemicals between the blood and nearby tissue via diffusion. Thus circulation paths consist of flow in larger vessels for most of the travel distance and a relatively short distance for branching into and out of capillaries.

\subsection{Flow at Vessel Branches}

\begin{figure}
\centering 
\begin{tabular}{cc}
\includegraphics[width=\mfigwidth]{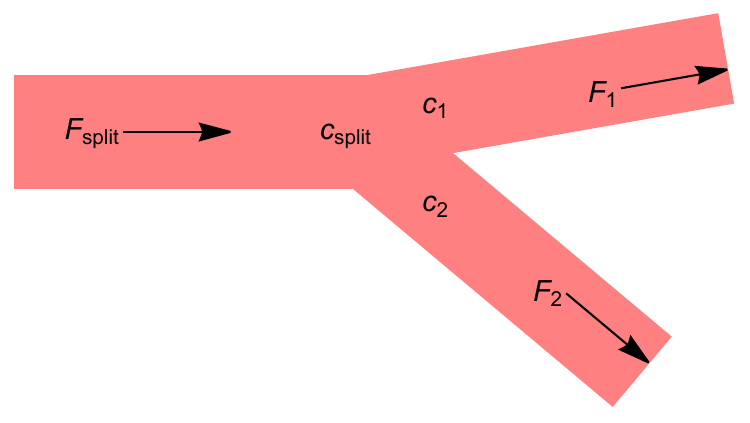}	&	\includegraphics[width=\mfigwidth]{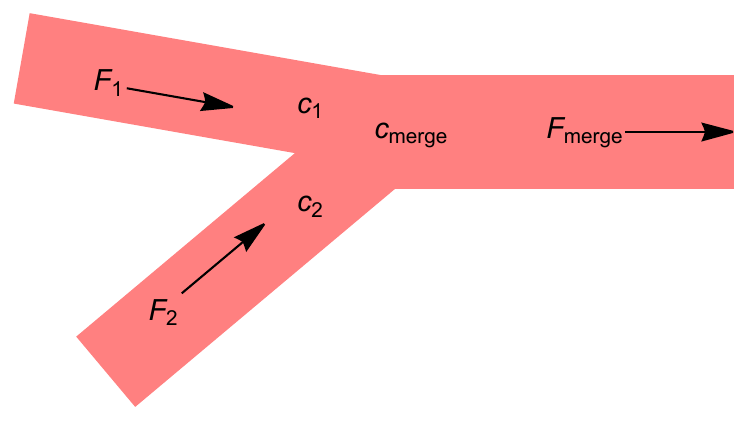} \\
(a) & (b) \\
\end{tabular}
\caption{Concentration $c$ and flow $F$ near branches. (a) A vessel splits into two branches. (b) Two vessels merge into one.}\figlabel{branching}
\end{figure}

The flow through different vessels combine where they join, as illustrated in \fig{branching}. Specifically, pressure is continuous at branches and total flow is conserved:
\begin{equation}\eqlabel{branch flow}
\begin{split}
P_{\textnormal{main}} &= P_1 = P_2 \\
F_{\textnormal{main}} &= F_1 + F_2 \\
\end{split}
\end{equation}
where $F_i$ is the flow rate in the $i$-th branch and $P_i$ is the pressure where that branch joins the main vessel. The subscript ``main'' denotes the main vessel, labeled ``split'' or ``merge'' for splitting or merging vessels, respectively, in \fig{branching}.

For a network of segments, \eq{branch flow} relates the pressure and flow of segments to each other and to the overall pressure drop and flow through the network. The values for individual segments combine in the same way as voltage and current in an electrical circuit.

\subsection{Concentration at Vessel Branches}

Blood carries oxygen in plasma and bound inside red cells. Within a segment, the change in concentration is determined in the same way as for the average circuit model~\cite{hogg21-average}. Evaluating concentration in a network of segments also requires identifying how concentration changes where segments join.

For splitting vessels, illustrated in \fig{branching}a, the concentration in the main vessel determines the concentration at the start of the branches. The concentrations are the same in the three vessels where they join:
\begin{equation}\eqlabel{branch split}
c_{\textnormal{split}} = c_1 = c_2
\end{equation}
where $c_i$ is the concentration in the $i$-th splitting branch just after the split.

Where vessels merge, oxygen concentration depends on how quickly oxygen from the merging vessels mixes as the blood flows into the main vessel. The aggregation of vessels into segments treats oxygen as well-mixed throughout the vessel cross section at each location along the segment. 
With this assumption, the concentration in the main vessel is the average of the concentration of the merging vessels, weighted by the flow in each one:
\begin{equation}\eqlabel{branch merge}
c_{\textnormal{merge}}  =  \frac{c_1 F_1 + c_2 F_2}{F_1 + F_2} 
\end{equation}
where $c_i$ is the concentration in the $i$-th merging branch just before the merge, and $F_i$ is the flow rate in that branch.
To see this, consider the flow into the merging vessel in a short time $\Delta t$. During this time, the volume of blood from the branches entering the large vessel is $(F_1 + F_2) \Delta t$. This blood carries $(c_1 F_1 + c_2 F_2) \Delta t$ oxygen molecules. The ratio of these quantities is the concentration at the start of the merged vessel given in \eq{branch merge}.
\sectA{mixing 3D} discusses the applicability of this mixing assumption.

\subsection{A Network of Vessel Segments}

\begin{figure}
\centering  
\begin{tabular}{cc}
\includegraphics[width=2.6in]{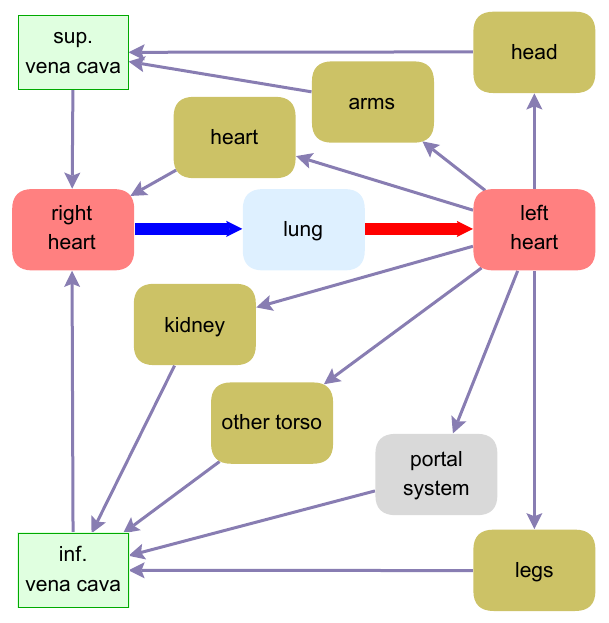}	&	\includegraphics[width=3.0in]{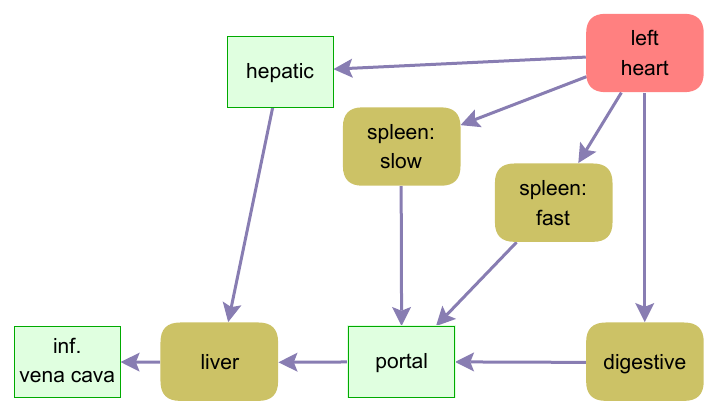}  \\
(a) & (b) \\
\end{tabular}
\caption{Graph of the circulation grouped by segments with similar properties. (a) Overall circuit. The large blue and red edges indicate the pulmonary circulation from the right heart through the lungs and to the left heart. The other edges are the systemic blood flow. (b) Detail of the flow through the portal system.}\figlabel{circulation}
\end{figure}

Aggregating vessels into segments with similar transit time, hematocrit profile and tissue demand is somewhat arbitrary.  For example, \eq{branch split} indicates that branches of a splitting vessel receive blood with the same concentration as in the main vessel, so there is no change in concentration due to splits. If such vessels also have similar hematocrit profiles, modeling them separately or as a single aggregated segment makes no difference. In particular, only small vessels have hematocrit substantially different from its overall value in the body, so moderate and large arteries need not be modeled as distinct segments. 
On the other hand, merging vessels combine blood with different oxygen concentration according to \eq{branch merge}. In particular, blood from paths with significantly different transit times combines in large veins. Since such paths can have different concentrations,  these veins must be distinct segments to account for how merges affect oxygen concentration.

These considerations lead to the circulation network in \fig{circulation} where boxes indicate the segments considered in the model. The arrows indicate the direction of blood flow between the segments.  
For example, the right and left chambers of the heart are separate pumps, which are both distinct from the coronary circulation through the heart muscle, even though all these flows are part of the heart considered as an organ. Conversely, the left and right lungs function as a single organ providing oxygen to the blood, and so are grouped into a single lung segment. Similarly, the two kidneys form a single segment.
\sectA{segments and vessels} describes how the segments in the network model relate to vessel structure and flows.

\section{Vessel Segment Parameters}\sectlabel{parameters}

\begin{table}
\centering
\begin{tabular}{lcc}
total blood volume		& \bloodVolume								& $5.4 \,\liter$ \\
total blood flow			& \bloodFlow								& $5\,\liter/\minute$ \\
average circulation time	& $\bloodAvgTransit = \bloodVolume/\bloodFlow$ 	& $65\,\second$ \\   
\hline
\multicolumn{3}{c}{\bf{pressure drop}}\\
systemic circuit		&  	&  $95 \,\mmHg$ \\
pulmonary circuit	&  	&  $8 \,\mmHg$ \\
\end{tabular}
\caption{Typical overall blood flow parameters for a person at rest. Pressure drops through the system and pulmonary circuits are averages over the heart cycle~\cite{feher17}.}\tbllabel{blood parameters}
\end{table}

In combination, the segments of the circulation network of \fig{circulation} carry the entire blood flow, with aggregate parameters given in \tbl{blood parameters}. 
This section gives the parameters for each segment in the network. These parameters determine how blood distributes among the segments. They also provide key quantities determining oxygen consumption in each segment: how long blood takes to move through the segment and the number of red cells available to replenish oxygen in the blood plasma, i.e., the hematocrit profile of the segment.

\subsection{Segment Flow Parameters}\sectlabel{flow parameters}

\begin{table}
\centering
\begin{tabular}{lccccc}
segment		& transit time	& resistance		& flow		& volume	& pressure drop \\ 
			& $\second$	& $\vresistance$ 	&  $\liter/\min$	& $\milliliter$	&  $\mmHg$ \\ \hline
\bf{heart and lungs} \\
 \text{left heart chambers} & 2.5 & \text{N/A} & 5 & 208 & \text{N/A} \\
 \text{right heart chambers} & 2.5 & \text{N/A} & 5 & 208 & \text{N/A} \\
 \text{heart (coronary)} & 3 & 3800 & 0.2 & 10 & 95 \\
 \text{lungs} & 6 & 12.8 & 5 & 500 & 8 \\
\bf{upper body} \\
\text{head} & 25 & 945 & 0.8 & 333 & 94.5 \\
 \text{arms} & 60 & 1890 & 0.4 & 400 & 94.5 \\
\bf{torso} \\
\text{spleen: fast} & 25 & 5210 & 0.13 & 56 & 87.8 \\
 \text{spleen: slow} & 750 & 46900 & 0.01 & 187 & 87.8 \\
 \text{digestive} & 45 & 868 & 0.81 & 607 & 87.8 \\
 \text{liver} & 25 & 33 & 1.2 & 500 & 4.9 \\
 \text{kidney} & 15 & 752 & 1 & 250 & 94 \\
 \text{other torso} & 51 & 940 & 0.8 & 680 & 94 \\
\text{hepatic} & 2 & 2970 & 0.24 & 8 & 89 \\
 \text{portal} & 3 & 10 & 0.96 & 48 & 1.2 \\
 \bf{lower body} \\
\text{legs} & 120 & 1250 & 0.6 & 1203 & 94 \\
\bf{major veins} \\
sup. vena cava		& 1 & 3.3 & 1.2 & 20 & 0.5 \\
inf. vena cava		& 3 & 2.2 & 3.6 & 180 & 1 \\
\end{tabular}
\caption{Circulation segment parameters: transit time  \tVessel, vascular resistance \RVessel, flow rate \FVessel,  blood volume \VVessel\ and pressure drop $\Delta P$. The values are averaged over the heart cycle. 
}\tbllabel{segment parameters}
\end{table}

The segment properties discussed in \sect{variation model} describe the flow through each segment as an aggregation over the vessels that form the segment. These flow parameters vary considerably among different people, times (e.g., during exercise or after a meal) and positions (e.g., standing pools more blood in the legs than a horizontal position). Moreover, different measurement techniques lead to additional variation.
Thus values for these properties have considerable variation.

As a specific case to evaluate robot power variation, \tbl{segment parameters} gives representative flow parameters for a person at rest.
These values arise from the estimated values for individual segments described below, with adjustments to satisfy relationships among the parameters. Specifically, \eq{vessel} gives two relations among the five flow parameters for each segment and \eq{branch flow} relates parameters in connected segments. 
In addition, the segment values must combine to match the properties of the overall circulation given in \tbl{blood parameters}. That is, the sum of the blood volumes in the segments must equal the total blood volume \bloodVolume, and total flow rate through the segments must equal \bloodFlow, which is the same for the pulmonary and systemic circulation.
In addition, the total pressure drops through the segments must match the total pressure drops of the pulmonary and systemic circulation.

The heart pumps the total blood flow, $\bloodFlow$ and contains about $400$ to $450\,\milliliter$ of blood. Each beat ejects about 60\% of the ventricle blood volume~\cite{feher17}.
The coronary circulation, i.e., through the heart segment of \fig{circulation}, uses about $5\%$ of the heart output~\cite{feher17}.
These values are the basis for the parameters of the heart chambers and coronary circulation in \tbl{segment parameters}.

Transit through a lung takes about $5\,\second$ and the lungs contain about $450\,\milliliter$ of blood~\cite{dock61}. Adding time and volume for the pulmonary artery and vein included in the lung segment of \fig{circulation} gives the value for lungs in \tbl{segment parameters}.
The lung capillaries contain about $70\,\milliliter$ of blood, which passes through a capillary network, i.e., from the end of arterial tree to the beginning of venous tree, in about $0.75\,\second$~\cite{feher17}.

The circulation model of \fig{circulation} includes large veins because they mix blood from paths with significantly different transit times, and hence different oxygen consumption by robots. The mixing in these veins is an important source of variation compared to a single average circuit. The flow in these veins is the sum of flows from incoming vessels. 
In particular, the superior and inferior vena cava collect blood returning from the upper and lower body, respectively. These are large vessels with relatively rapid blood flows~\cite{freitas99}, low pressure drops and short transit times.

The upper body in \fig{circulation} consists of flow through the arms and head.
Flow to the brain is about $0.7\,\liter/\minute$~\cite{feher17}, which accounts for most of the flow to the head.
The brain contains about $150\,\milliliter$ of blood~\cite{sourbron09}. Adding blood in the rest of the head and a portion of the volume of large arteries from the heart to head, not separately modeled in \fig{circulation}, gives the blood volume estimate for the head in \tbl{segment parameters}. 
At rest, flow to the arms is about half the flow to the brain~\cite{ahlborg91}.
The pressure drops through the head and arms are nearly equal to the total pressure drop of the systemic circuit since the superior vena cava has only a small change in pressure. From these values, \eq{vessel} determines transit time and vascular resistance for circulation through the head and arms.

The lower body consists of the legs, with the longest circulation distances in the body. Reported flow rate~\cite{holland98} and blood volume~\cite{nylin47} indicate transit times in the range of $1.5\mbox{-}2.5\,\minute$. As a specific value, \tbl{segment parameters} uses an intermediate value for the transit time.

The torso segments of \fig{circulation} consists of the portal system, flowing through digestive organs and the liver, and the other organs, muscles and skin of the torso. 

The perfusion of the liver is $1.2\,\liter/\minute$~\cite{feher17},   
which includes blood from both the hepatic artery and the portal vein carrying blood from digestive organs. These flows combine in the liver's capillaries~\cite{lautt09}. 
Thus the hepatic and portal segments of \fig{circulation} include not just the hepatic artery and portal vein themselves, but also branches from those vessels in the liver until they meet in liver capillaries. This means these segments include some of the blood transit through the liver, while the liver segment has the rest, i.e., the flow starting from liver capillaries.
The portal vein provides about $80\%$ of the liver's blood and the hepatic artery provides the rest~\cite{freitas99}. 
This split gives the flows for the hepatic and portal segments in \tbl{segment parameters}.
The liver has much lower vascular resistance than other organs~\cite{lautt09}.

The total flow through the two kidneys is $1\,\liter/\minute$~\cite{feher17}   
and they contain about $70\,\milliliter$ of blood~\cite{effros67}. 
From \eq{vessel}, these values correspond to about $4\,\second$ transit through the kidney.
Adding time to reach the kidney from the heart and return to the vena cava, as illustrated in \fig{circulation}, gives the estimates for the kidney vessel segment in \tbl{segment parameters}.

The amount of blood stored in the spleen, its transit time and details of the flow vary considerably~\cite{groom02,witte83}. About $5\%$ of cardiac output goes to the spleen, of which 10\% passes through narrow slits, leading to long transit times~\cite{cataldi17}.
Since transit time is an important factor for oxygen consumption, \fig{circulation} includes two segments for the spleen: ``fast'' for the majority of the flow that avoids the slits and ``slow'' for the rest.
\tbl{segment parameters} uses representative values for these flows. 

The ``other torso'' vessel segment in \fig{circulation} accounts for the remainder of the total blood flow \bloodFlow. Its blood volume is chosen to combine with the segment volumes to equal the body's total blood volume, \bloodVolume. These values of flow and volume give the average transit time via \eq{vessel}.

The graph of vessel segments in \fig{circulation} has ten paths around the systemic circulation, i.e. flow starting with leaving the left heart and ending with its return to the right heart. All these paths share the same pulmonary circuit. 
As a summary of \tbl{segment parameters}, \fig{paths} shows the flow rates and transit times for each of these paths. The sum of the flow and blood volume in these paths equals \bloodFlow\  and \bloodVolume, respectively, as given in \tbl{blood parameters}.

\begin{figure}
\centering 
\includegraphics[width=\widefigwidth]{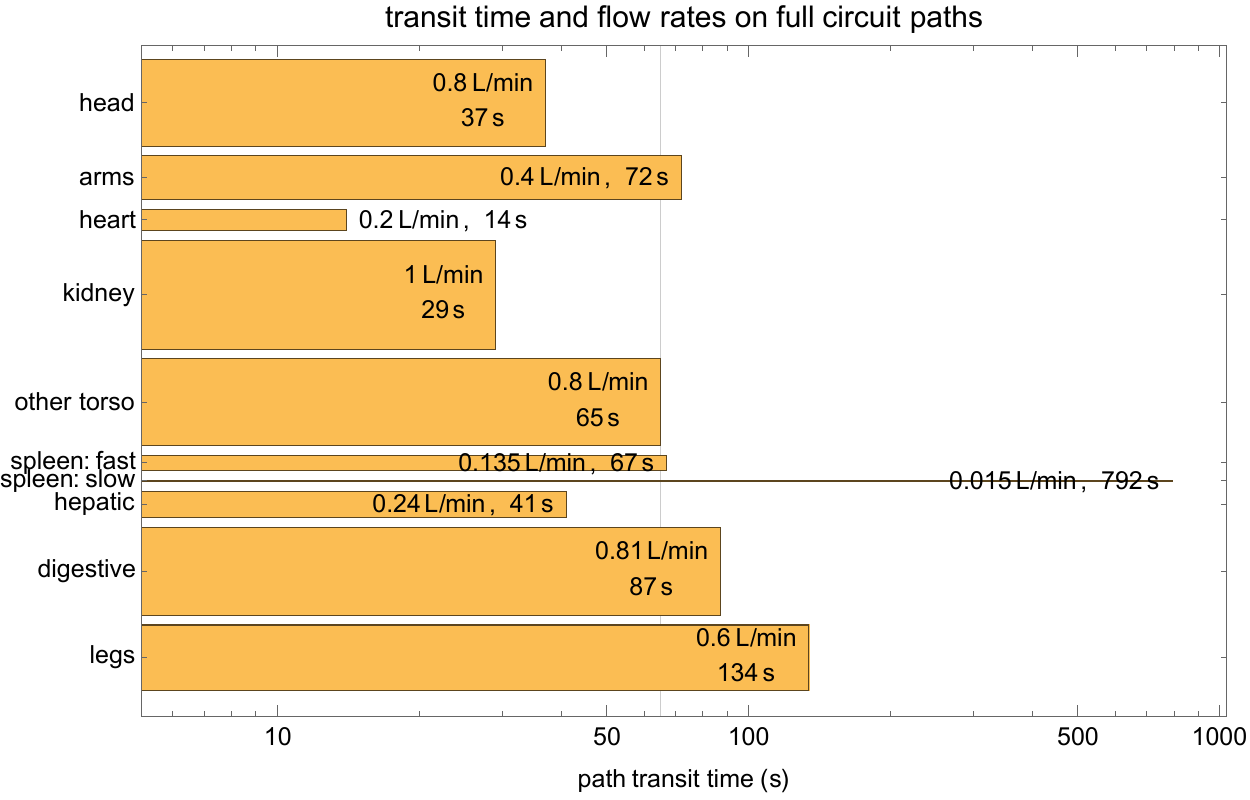} 
\caption{Path flow and total transit time (on a log scale) indicated by the height and width of each bar, respectively. Each path is a full loop through both the systemic and pulmonary circulation. 
Path labels are the name of the first node along the path after leaving the heart in the graph shown in \fig{circulation}.
The total flow of these paths and their flow-weighted average transit time equal \bloodFlow\ and \bloodAvgTransit, respectively, given in \tbl{blood parameters}.
}\figlabel{paths}
\end{figure}

\subsection{Segment Hematocrit Profiles}\sectlabel{hematocrit profiles}

\begin{figure}
\centering  
\begin{tabular}{cc}
\includegraphics[width=2.6in]{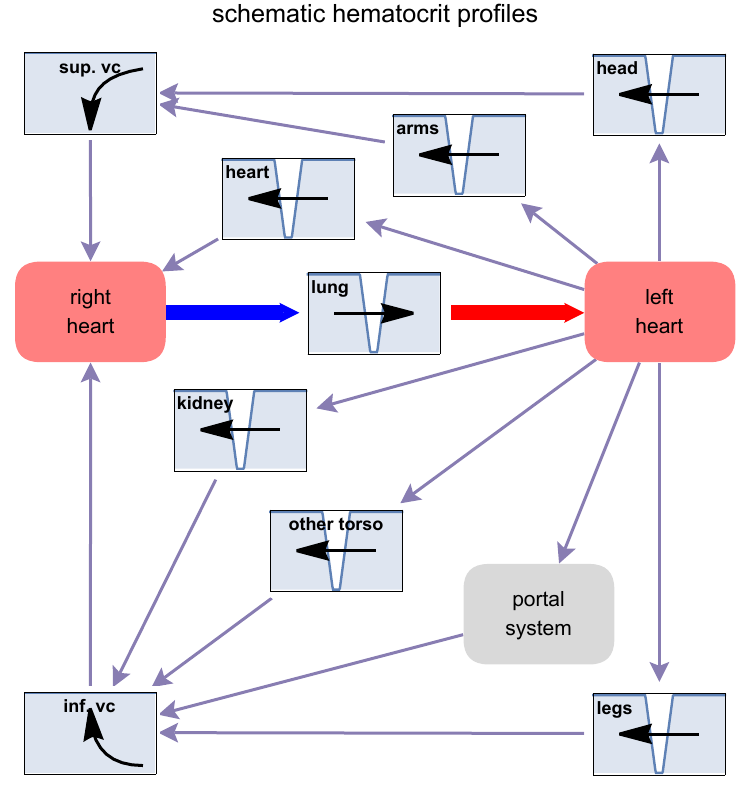}	&	\includegraphics[width=3.0in]{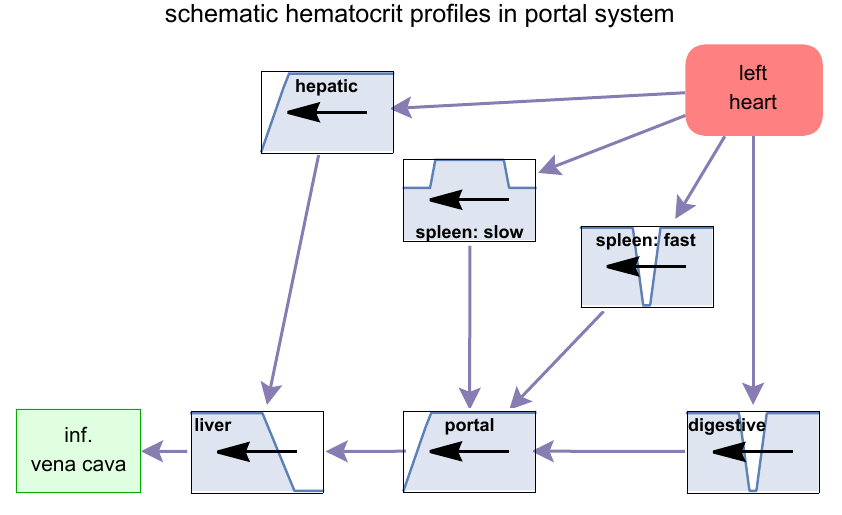}  \\
(a) & (b) \\
\end{tabular}
\caption{Schematic hematocrit profiles associated with the segments of the circulation model. Values range from $0.45$ in large vessels down to $0.33$ in capillaries, except for flow through the slow spleen compartment where hematocrit in capillaries is $0.71$. For clarity, the profiles show an exaggerated fraction of time in small vessels. Actual transit time through capillaries is one second, except for an extended transit through the slow compartment of the spleen. The arrows indicate the direction of flow. (a) Overall circuit. (b) Detail of the portal system.}\figlabel{hematocrit}
\end{figure}

Each segment of \fig{circulation} has an associated hematocrit profile, as illustrated schematically in \fig{hematocrit}. 
Each hematocrit profile is defined by a piecewise linear function that uses the typical hematocrit values in large vessels and capillaries described in \sect{hematocrit variation}. A linear interpolation between these values gives the hematocrit in intermediate-sized vessels, extended over the typical time required to pass through small branching vessels to and from a capillary estimated from lung branching~\cite{huang96}.
The precise form of this interpolation makes little difference in model behavior because blood spends relatively little time in small arteries and veins so the hematocrit in branching vessels has little effect on oxygen concentration.

Most segments consist of large vessels branching to capillaries and then converging to large vessels: the profiles for these segments consist of a drop in hematocrit from 0.45 to 0.33 near the middle of the segment. The hepatic and portal vein segments end just as blood mixes in liver capillaries. Thus the hematocrit profiles of these two segments drop toward the end of the segment, representing the decreasing hematocrit from large vessels to capillaries. Blood flow in large veins and the chambers of the heart does not pass through capillaries, so hematocrit does not change in those segments.

As discussed in \sect{hematocrit variation}, small vessels in the spleen's slow compartment have large hematocrit values. As a specific choice, the profile for this segment uses a hematocrit value, $0.71$, giving three times higher concentration of cells relative to plasma than their overall concentration in the body~\cite{wadenvik88}. This increased hematocrit applies to most of the transit time through this spleen compartment, i.e., as blood passes through narrow slits.

\subsection{Tissue Power Demand Parameters}

As described in \sect{tissue variation}, for the most part, variation in resting tissue demand is relatively small on the per capillary basis relevant for this model.
This allows estimating tissue power demand in the same manner as for the average circulation model~\cite{hogg21-average}: a cylinder of tissue, of radius $40\,\micron$, surrounds each capillary~\cite{krogh19,popel89} and has the average resting power demand for tissue: $4\,\kilowatt/\meter^3$~\cite{freitas99}.
The radius of this tissue cylinder corresponds to a typical maximum distance of tissue supplied from a capillary, which is a few cell diameters.
For a typical capillary length of $1\,\millimeter$, this corresponds to tissue power consumption of about $20,000\,\picowatt$ per capillary.
By comparison, on average, the number density of capillaries in tissue is $600/\millimeter^3$~\cite{freitas99} so average resting tissue demand corresponds to about $7000\,\picowatt$ per capillary. 
Thus the model uses about three times the average tissue demand per capillary, providing some accommodation for tissues with higher demand per capillary.

The two exceptional cases discussed in \sect{tissue variation} require modifying the tissue power demand.
For the first exception, the model sets tissue demand in the coronary capillaries so that, with no robots, the oxygen concentration after passing through a capillary is reduced to $30\%$ of its incoming value. Increasing tissue power demand to $50\,\kilowatt/\meter^3$ achieves this behavior with the tissue cylinder geometry used in the model~\cite{hogg21-average}.
For the second exception, in the slow moving blood in the spleen, the power demand is set to give the same total demand during that transit as occurs during the one-second capillary transit in typical resting tissue.  
This is accomplished by decreasing the tissue power per capillary by the ratio of capillary transit time in this part of the spleen (given in \tbl{segment parameters}) compared to one second.

\section{Robot Power and Oxygen Concentration}\sectlabel{robot power}

The model evaluates oxygen concentration in each vessel segment as it changes due to robot and tissue consumption as well as release from blood cells. This uses the transit time and hematocrit profile of the segment. Tissue consumption only occurs in capillaries. On the time scales considered here, oxygen release from blood cells is rapid~\cite{clark85}, so the model assumes oxygen in the cells remains in equilibrium with the concentration in the surrounding blood plasma, as described by the Hill equation~\cite{popel89}. 
Oxygen in blood plasma and cells is replenished in lung capillaries.

The model assumes robots flow at the same speed as cells.
Robots are taken to be spheres to determine the rate oxygen reaches their surfaces via diffusion~\cite{berg93} depending on its concentration in the surrounding blood plasma.

To compare this network model with the average circulation model~\cite{hogg21-average}, the same robot and oxygen consumption properties~\cite[Appendix D]{hogg21-average} are used. In particular, this considers robots with radius $\rRobot = 1\,\micron$.
With these properties and the parameters of \sect{parameters}, the network model determines oxygen concentration and robot power throughout each segment of \fig{circulation}. 

The average circulation model showed complete oxygen depletion toward the end of the circulation loop, i.e., in large veins, when using a trillion robots that consume all oxygen diffusing to their surfaces~\cite{hogg21-average}. On the other hand, $10^{11}$ robots gave significant but not total oxygen reduction. Thus this is an important range of robot numbers to evaluate with the network model in this paper.  
This evaluation shows that merging flows from paths of differing lengths result in somewhat higher oxygen concentration in large veins than suggested by the average-circuit model. Instead, the lowest concentrations occurs in long circulation loops just before they merge with blood from shorter circuits.

\begin{figure}
\centering 
\includegraphics[width=\widefigwidth]{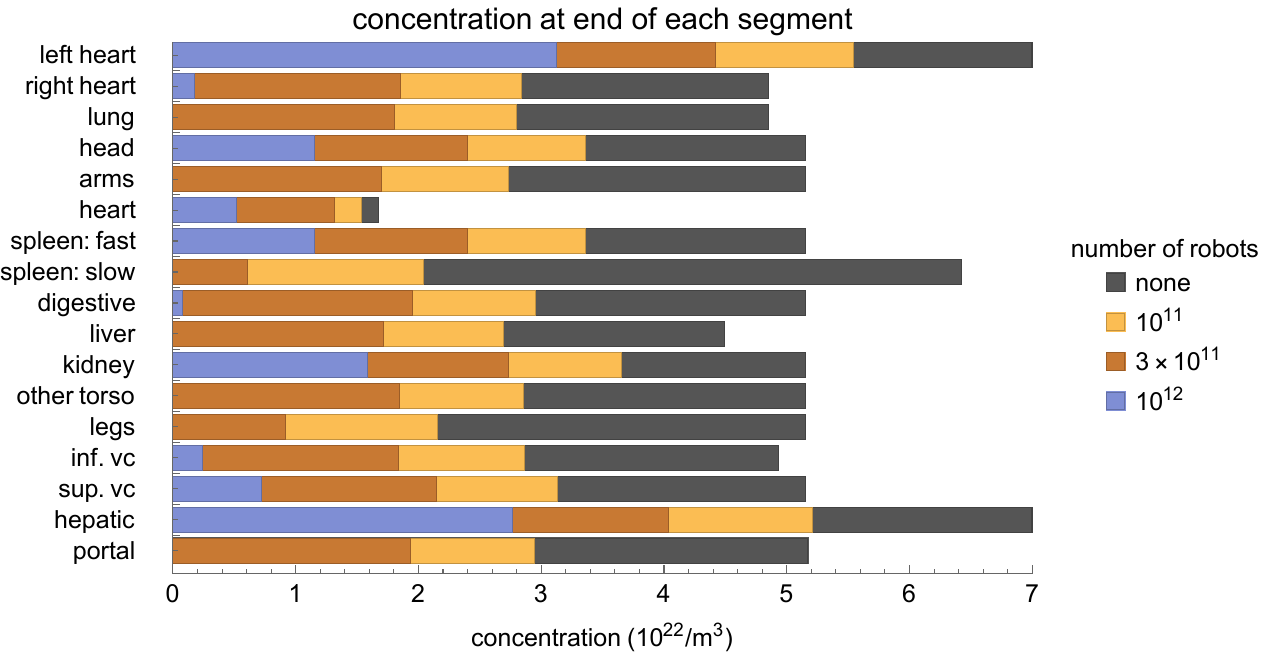}
\caption{Oxygen concentration at the end of each vessel segment for various numbers of robots in the blood.}\figlabel{segment concentration}
\end{figure}

The concentration of oxygen decreases as robots and tissue consume it, so the lowest concentration in each segment occurs at the end of that segment.
\fig{segment concentration} shows the oxygen concentration at the end of each segment of \fig{circulation}. In most cases, the end of a segment occurs in a medium or large vein at some distance after the blood flows through a capillary. 
However, the left heart and hepatic segments have no capillary passage during or prior to that segment, so concentration at the end of those segments is the same as that in lung capillaries when there are no robots. The other segments have some reduction in oxygen concentration even with no robots, due to consumption by tissue in capillaries.

For segments with long transit times, i.e., the legs and slow spleen segments, the concentration at the end of the segment is lower than that in the large veins where flow from those segments mixes with blood on faster paths. For example, $10^{12}$ robots completely deplete oxygen in these long paths while still leaving some oxygen in the vena cava. 

The location of the smallest concentration varies with the number of robots. With fewer robots, tissue consumption is relatively more important, e.g., leading to lowest concentration at the end of the heart segment. With large numbers of robots, lowest concentration occurs at the end of long segments, such as the legs and slow spleen segments.

\begin{figure}
\centering 
\includegraphics[width=\widefigwidth]{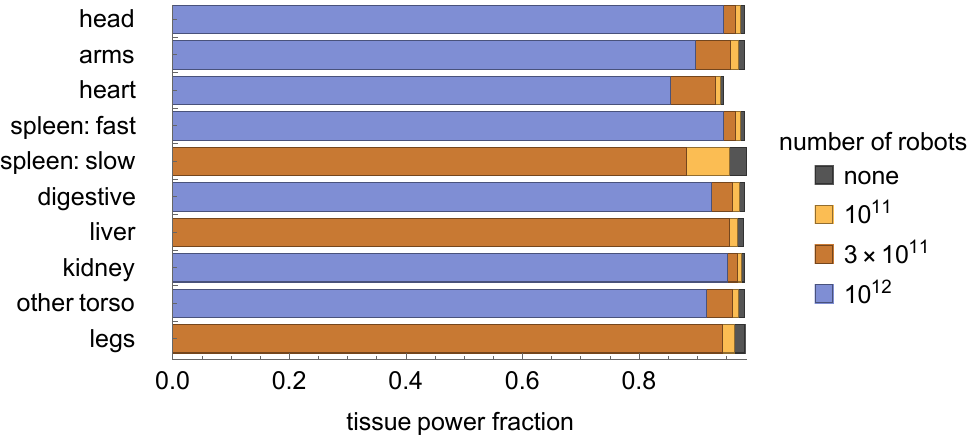}
\caption{Relative tissue power at the venous end of a body capillary, in the segments that include passage through a capillary, for various numbers of robots in the blood.}\figlabel{tissue power}
\end{figure}

A major consequence of oxygen reduction by robots is the possibility of depriving tissue of sufficient oxygen to support metabolic processes. Tissue consumes oxygen from capillaries, so a measure of tissue oxygen supply is the oxygen concentration at the end of a capillary rather than the end of vessel segments which, in many cases, are in veins.
Moreover, capillaries normally have much more oxygen than tissue requires at its basal metabolic rate~\cite{hogg10}.
For these reasons, the oxygen reductions shown in \fig{segment concentration} lead to relatively less reduction in tissue metabolism.

To quantify this behavior, \fig{tissue power} shows the tissue power at the venous end of a capillary relative to its maximum resting power demand, using the same tissue consumption model as for the average circulation study~\cite{hogg21-average}.
The relative tissue power is fairly close to one, and to the values when there are no robots in most of these segments even with $10^{12}$ robots. This figure shows where in the body $10^{12}$ robots would be particularly detrimental to tissue and therefore where to focus mitigations discussed in \sect{mitigation}. In particular, major reductions occur in the legs, liver and slow spleen segments.

\begin{figure}
\centering 
\begin{tabular}{cc}
\includegraphics[width=2.6in]{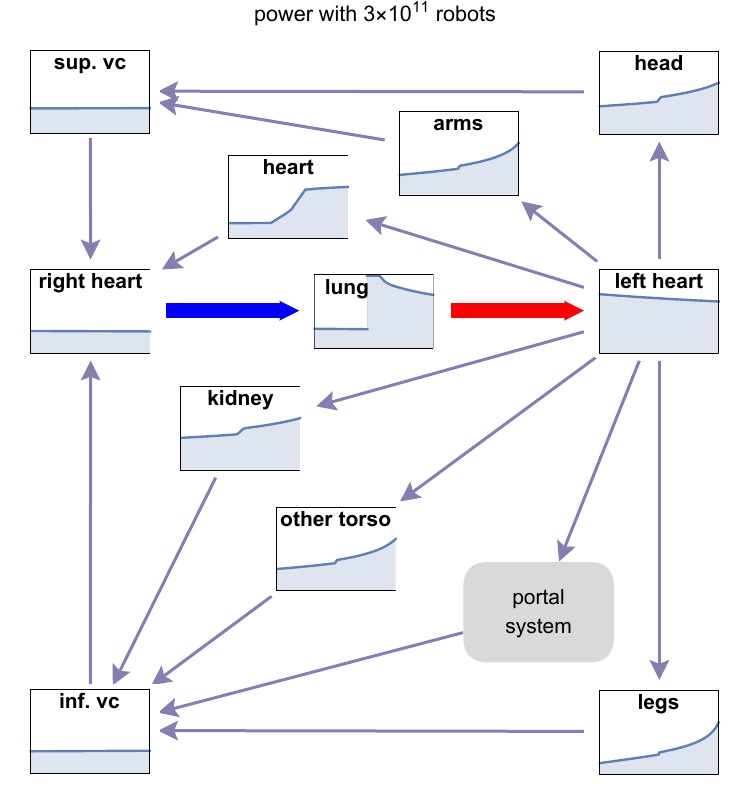}	&	\includegraphics[width=3.0in]{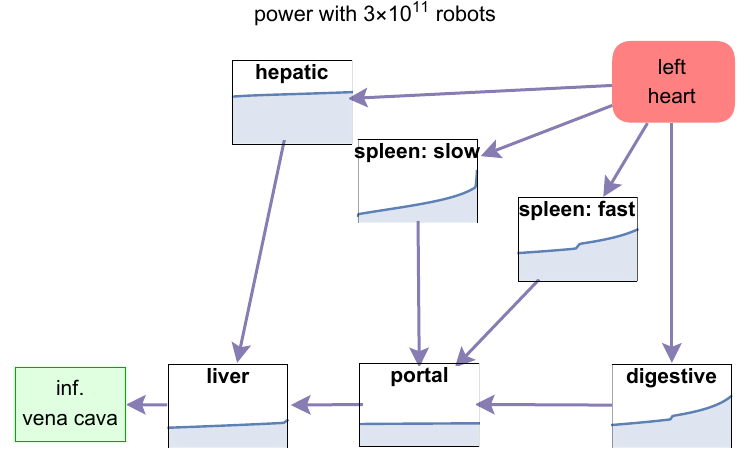}  \\
(a) & (b) \\
\end{tabular}
\caption{Robot power as a function of time during passage through each vessel segment with $3\times 10^{11}$ robots in the blood. 
Each plot scales the time so the robot transit time in its segment corresponds to the same width. The vertical axis of each plot shows power in the range of $0$ to $600\,\picowatt$. The orientation of each plot corresponds to the direction of the flow, indicated by the arrows and is the same as in \fig{hematocrit}. (a) Overall circuit. (b) Detail of the portal system.}\figlabel{robot power}
\end{figure}

The above discussion focuses on how robots affect oxygen concentration at the end of each segment and in body capillaries.
Oxygen concentration changes from the start to the end of each segment, with a corresponding change in the power available to robots.
\fig{robot power} is one example of how power available to the robots varies as they move through each segment. Power is largest in lung capillaries and lowest, about $70\,\picowatt$, just before blood returning from the legs merges with flow from the torso and at the end of the slow transit through the spleen.

\section{Example Behaviors Along Circulation Paths}\sectlabel{some circulation paths}

Paths through circulation have considerable variation in transit time (see \fig{paths}) and hematocrit profiles (see \fig{hematocrit}). This section describes three of these paths: the portal flow to the liver, flow in the spleen where capillary hematocrit is enhanced, rather than reduced, and the coronary circulation.

\subsection{Portal Flow}\sectlabel{portal}

\begin{figure}
\centering 
\includegraphics[width=\figwidth]{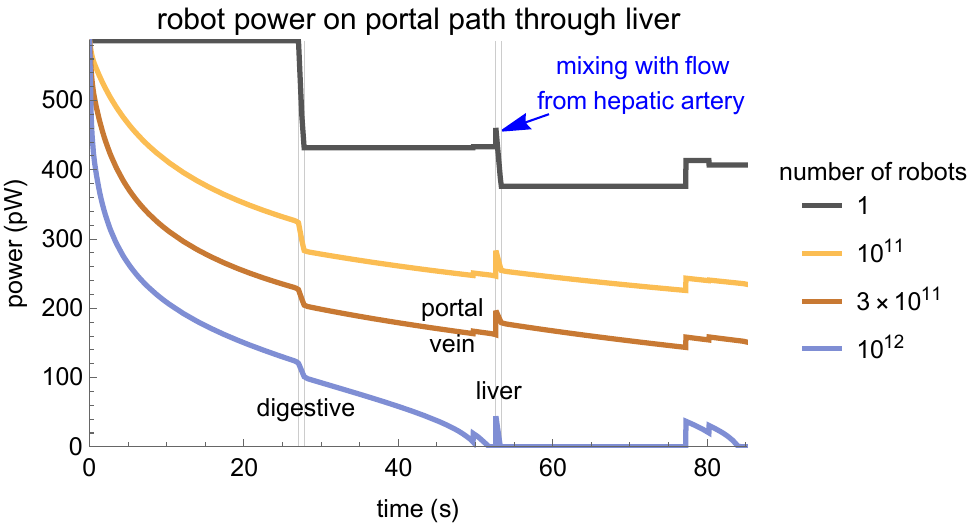}
\caption{Power for robots moving with portal flow for various numbers of robots in the blood. The blue arrow indicates the increased power due to mixing with blood from the hepatic artery in a liver capillary. The two pairs of light gray vertical lines indicate when the robot passes through a capillary. Labels near the lower curves indicate when the robot passes through a digestive organ, the portal vein and the liver.}\figlabel{portal power}
\end{figure}

\fig{portal power} shows the power available to robots traveling through the portal system. The figure shows the two capillary transits on this path: first in a digestive organ and then in the liver. 
Robots entering the second capillary have a brief increase in power due to the mixing with arterial blood. A robot could use these changes to determine when it passes through a second capillary in a circuit, i.e., is in blood that is part of a portal flow, as a navigation aid in combination with other organ specific properties~\cite{freitas99}. 

The top curve in \fig{portal power} is for a single robot. Robot missions will involve many more robots, so this curve is an indication of the power available to a few robots for burst use if the vast majority of robots choose to consume much less oxygen than reaches their surfaces, as discussed as a mitigation strategy for the average circulation model~\cite{hogg21-average}. In this case, most of the oxygen consumption is due to tissue metabolic demand during the two capillary transits of the portal circuit.

The lowest curve in \fig{portal power}  shows $10^{12}$ robots deplete oxygen in the blood entering larger veins in the liver. The concentration, and hence robot power, increase as that flow from the liver merges with blood from other organs.

By contrast with the flow through the portal vein, robots entering a liver capillary directly from the hepatic artery, i.e., without traveling through the portal circulation, will experience a reduction in power as they mix with blood from the portal vein. The mixture of robots with these distinct histories, i.e., about 80\% traveling via the portal flow and the rest directly from the hepatic artery, provides another indication of portal flow the robots could identify by communicating with nearby robots in the liver. This communication would allow robots to compare measurements along the two paths to the liver capillaries, thereby supplementing their individual measurements. More generally, robots from blood mixing in larger veins will contain robots returning from different paths. Neighboring robots that compare their recent measurements could gain a global perspective on their individual sensor values, which may help identify anomalous values.

\subsection{Flow in the Spleen}\sectlabel{spleen}

\begin{figure}
\centering 
\includegraphics[width=\figwidth]{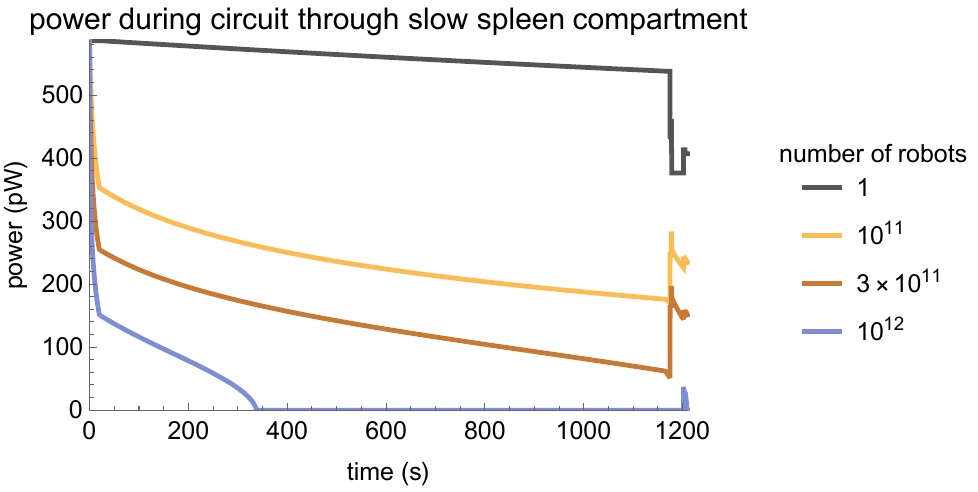}
\caption{Robot power during transit of slow spleen compartment for various numbers of robots in the blood.}\figlabel{spleen power}
\end{figure}

The capillaries of the slow flow through the spleen are unusual in having higher, rather than lower, hematocrit, as discussed in \sect{hematocrit variation}.
This model assumes robots move with the same speed as cells, so the increased hematocrit corresponds to robots taking longer to transit the capillary. Thus there are competing effects on the oxygen available to robots. 
On the one hand, higher hematocrit gives more cells to replenish oxygen consumed by robots while in capillary. On the other hand, the slower flow means cells and robots spend more time in the capillaries thereby consuming oxygen for a longer time before returning to larger vessels and, eventually, to the lung. 

\fig{spleen power} shows the consequences of these effects on robot power as they pass through the slow spleen compartment.
This shows the long time required to pass through the spleen dominates the increase in cells. In particular, robots consuming oxygen as rapidly as it diffuses to their surfaces completely deplete oxygen when there are $10^{12}$ robots in the circulation and blood cells are completely emptied of their oxygen. 
These desaturated cells leaving the spleen do not contribute oxygen as they merge with other blood flowing into the liver through the portal vein. This contributes to substantial reduction to oxygen available to the liver with this number of robots, as seen in \fig{tissue power}. 
However, the reduction in oxygen available to tissue in these organs is not as significant as in those where blood mainly provides support for tissue metabolism. This is because organs such as the spleen, liver and kidney mainly filter blood, so the flow through these organs is higher than needed to support the cells of the tissue~\cite{feher17}.

As a caveat on this discussion, the complex microcirculation of the spleen~\cite{groom02} may lead to more complex hematocrit profiles than considered here if they depend on path and transit time through the spleen. There is also the possibility that robots and cells move differently, contrary to the assumed behavior in this model. This possibility is because micron-size robots are smaller and stiffer than cells but larger than most nanoparticle drugs, in particular larger than the limit of about $200\,\nanometer$ of particle size able to pass through the slits in the spleen~\cite{cataldi17}. This may lead to robots taking somewhat longer to transit the spleen than cells and hence a larger concentration of robots in that portion of the blood, which could lead to even lower oxygen concentration. Alternatively, robots with locomotion capability could actively adjust their transit, e.g., to reduce the number flowing into the slow flow of the spleen.

\subsection{Coronary Circulation}\sectlabel{coronary}

\begin{figure}
\centering 
\includegraphics[width=\figwidth]{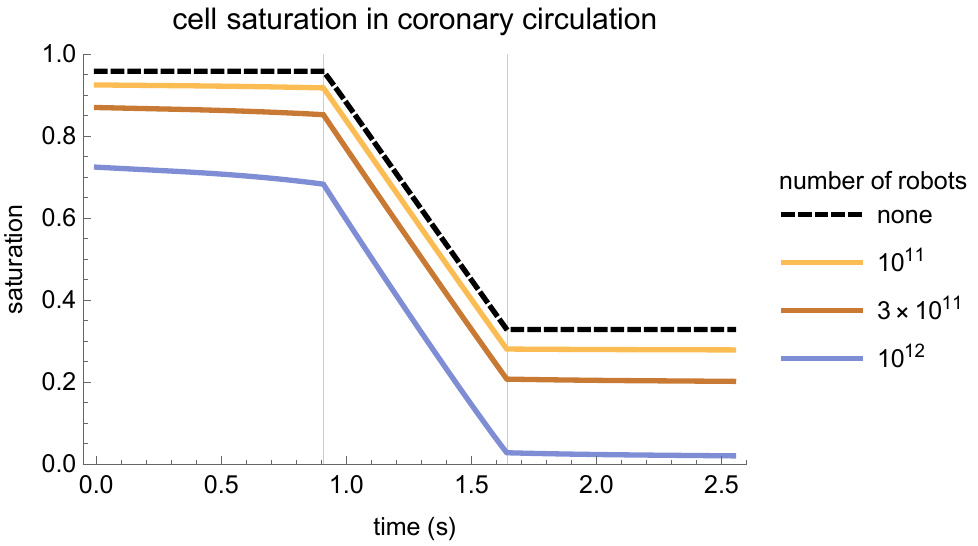}
\caption{Cell saturation in coronary circulation for various numbers of robots in the blood. The light gray vertical lines indicate when the flow is in a capillary.}\figlabel{coronary saturation}
\end{figure}

As described in \sect{tissue variation}, the large tissue demand and short transit time of the coronary circulation mean that tissue accounts for a significant portion of the oxygen use during this circuit, even with substantial numbers of robots. 
This is the opposite of the situation in longer circulation loops, where these numbers of robots consume far more of the oxygen than does tissue~\cite{hogg21-average}. \fig{coronary saturation} confirms this behavior in terms of how robots affect cell saturation.

\begin{figure}
\centering 
\includegraphics[width=\figwidth]{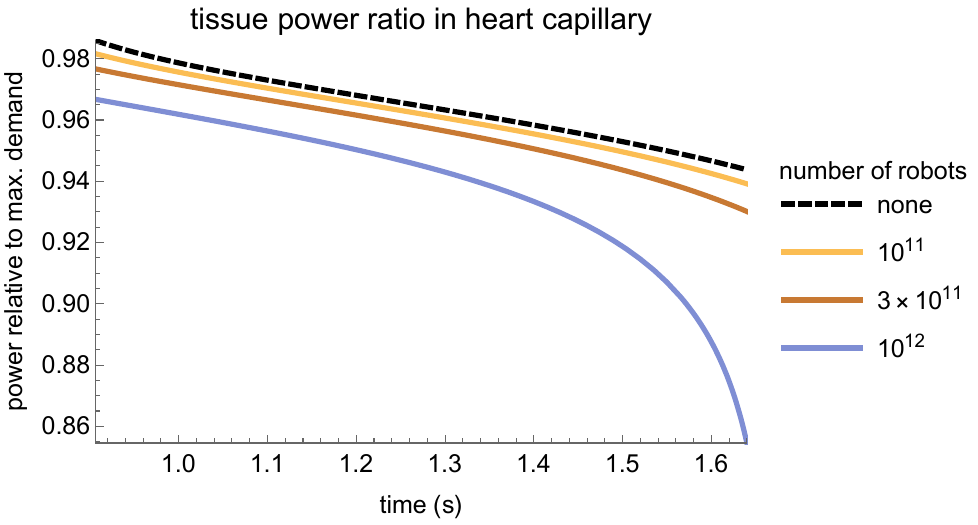}
\caption{Relative tissue power in coronary capillaries for various numbers of robots in the blood. The horizontal axis indicates time from the start of the coronary circulation, as shown in \fig{coronary saturation}, but only includes the time during which tissue consumes oxygen, i.e., while blood passes through a capillary.}\figlabel{coronary tissue power}
\end{figure}

Since tissue consumes so much of the available oxygen, even at its resting metabolic demand, an important issue is the extent to which robot consumption lowers the oxygen enough to reduce tissue power. \fig{coronary tissue power} shows the effect on tissue power of robots passing through coronary capillaries. The reduction is relatively small even for $10^{12}$ robots, for which the minimum relative tissue power is about $86\%$. This is primarily due to the short transit time of this circuit: diffusion-limited robots do not spend enough time in the coronary circulation to significantly reduce tissue power. However, even relatively small increases in transit time through coronary circulation, e.g., a few seconds, could lead to much more significant reductions. 

This sensitivity to capillary transit time contrasts with circulation loops taking tens of seconds, much longer than the one-second typical capillary transit time. In those cases, robot oxygen consumption mainly occurs in larger vessels so even relatively large variations in capillary transit or tissue consumption have relatively little effect on overall oxygen concentration with the number robots considered here.

\section{Sensitivity to Circulation Parameter Variation}\sectlabel{parameter sensitivity}

The circulation parameter values in \tbl{segment parameters} are representative of the differences among circulation paths through the body. However, there is considerable uncertainty in these values. This section evaluates how variation in these parameters affect robot power and oxygen concentration.

\subsection{Constrained Parameter Variation}

To evaluate the model's sensitivity to parameter variation, we consider variations in the circulation parameters that are consistent with well-established properties of the circulation. For this analysis, we focus on variation in systemic circulation for a person at rest. This indicates the effect of uncertainties or patient-specific variation in the values of \tbl{segment parameters} rather than variations due to changing activity levels or health status. 
Specifically, the parameters must match the total blood volume, flow rate and pressure change  given in \tbl{blood parameters}. We constrain the transit time and resistance parameters of the segments to be within a factor of two of the values in \tbl{segment parameters}.
 
For consistency with measured flows to individual organs~\cite{feher17}, we constrain parameters to give the following flows:  $0.2\,\liter/\min$ for the coronary circuit (i.e., the heart segment), $0.8\,\liter/\min$ through the head,  $1\,\liter/\min$ through the kidneys, and  $1.2\,\liter/\min$ through the liver, of which $20\%$ is through the hepatic artery. That is, the flows through these segments do not vary from the values in  \tbl{segment parameters}.

Other segments of the circulation have significant variation in flow. For these we apply looser constraints to maintain realistic circulation properties. Specifically, the arms have at least $0.2\,\liter/\min$, the legs have at least $50\%$ more than the arms, up to at most $1\,\liter/\min$ through the legs. The spleen receives between $0.1\,\liter/\min$ and $0.25\,\liter/\min$, with $90\%$ of that flow through the spleen's fast compartment.

To evaluate parameter sensitivity, 100 random samples of circulation parameters satisfying these constraints were generated. For each sample, the model was evaluated for the numbers of robots discussed above.

 \subsection{Consequences of Parameter Variation}

\begin{figure}
\centering 
\includegraphics[width=4.0in]{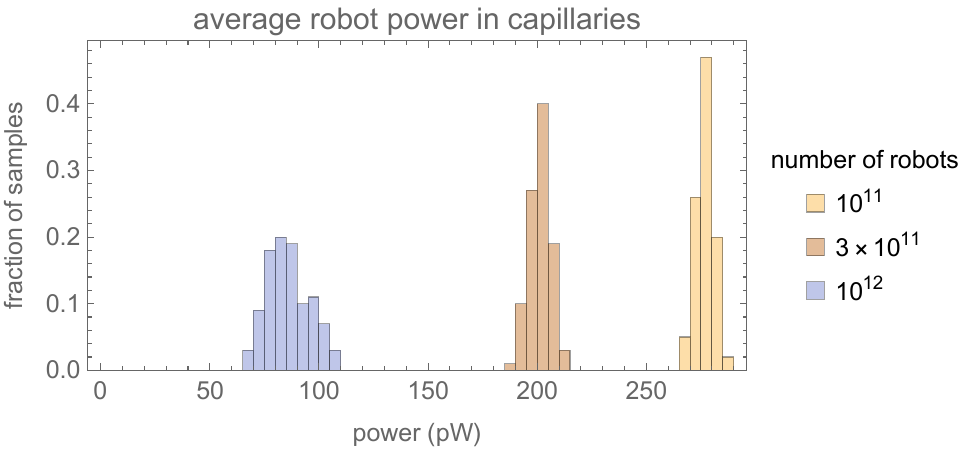}
\caption{Distribution of average power available to a robot when it reaches the end of a capillary of the systemic circulation. The average is weighted by the flow through each segment.}\figlabel{sensitivity: average power}
\end{figure}

With respect to average behavior, \fig{sensitivity: average power} shows the distribution in average power for the robots as they reach the end of a transit through a capillary in each of the segments of the systemic circulation shown in \fig{circulation}. 
This average is weighted according to the proportion of blood flow in each segment. For instance, with the constraints on parameter values described above, kidneys receive $20\%$ of the flow. The figure shows average power is fairly insensitive to variation in the circulation parameters.

For robots whose greatest power requirements occur as they pass through capillaries anywhere in the body, e.g., to measure or act upon nearby tissue cells, \fig{sensitivity: average power} indicates determining the average available power does not require accurate values for the circulation parameters of \tbl{segment parameters}. On the other hand, if robots must have adequate power everywhere in the body, the minimum power or oxygen concentration is more relevant than the average.

 \begin{figure}
\centering 
\includegraphics[width=\figwidth]{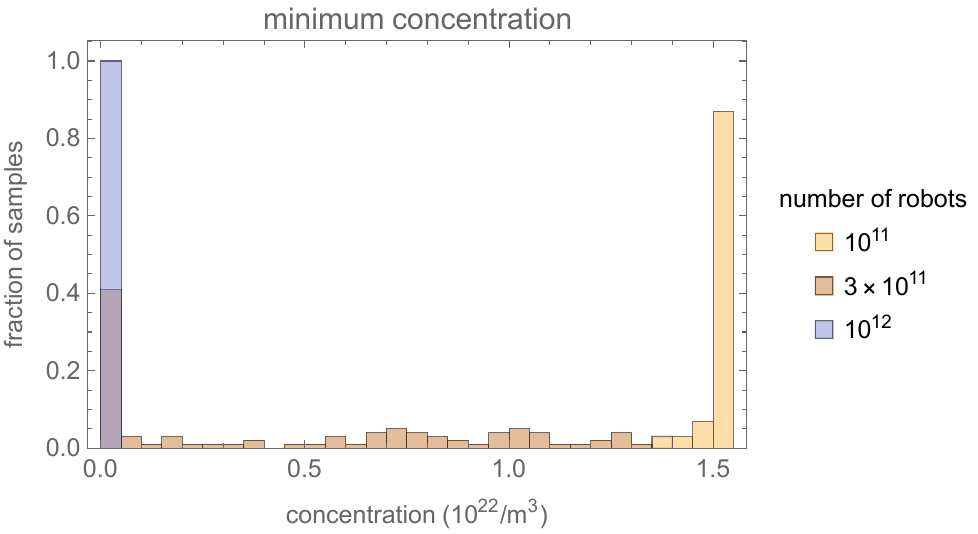}
\caption{Distribution of minimum oxygen concentration in vessel segments of the systemic circulation.}\figlabel{sensitivity: concentration}
\end{figure}

In each segment of the model, concentration monotonically decreases as blood passes through the segment, so the minimum in a segment occurs at its end, just before it merges with another segment, as described in \sect{robot power}. These minimum values vary among the segments, as shown in \fig{segment concentration} for the parameters of \tbl{segment parameters}. The smallest of these segment concentrations gives the minimum concentration in the body.
\fig{sensitivity: concentration} shows how this minimum varies with the circulation parameters.
This variation has little effect on the minimum concentration for $10^{11}$ or $10^{12}$ robots. However, with an intermediate number of robots, the minimum is highly sensitive to parameter variations: the figure shows concentrations for $3\times 10^{11}$ robots with different parameters cover nearly the full range of the histogram, with a significant fraction of the cases having very low concentration.

As with the behavior shown in \fig{segment concentration} for the parameters of \tbl{segment parameters}, in most cases, the minimum concentration occurs in the heart segment for $10^{11}$ robots and in the slow spleen segment for the other cases. Moreover, for $3\times 10^{11}$ robots, segments other than the slow spleen have concentrations similar to those shown in  \fig{segment concentration}.
Thus the large variation in minimum concentration for $3\times 10^{11}$ robots arises mainly from variation in transit time through the slow spleen segment. This means the variation in minimum concentration seen in \fig{segment concentration} could be significantly altered if robots actively avoid passing through the slow spleen segment~\cite{freitas03}.

\begin{figure}
\centering 
\includegraphics[width=\figwidth]{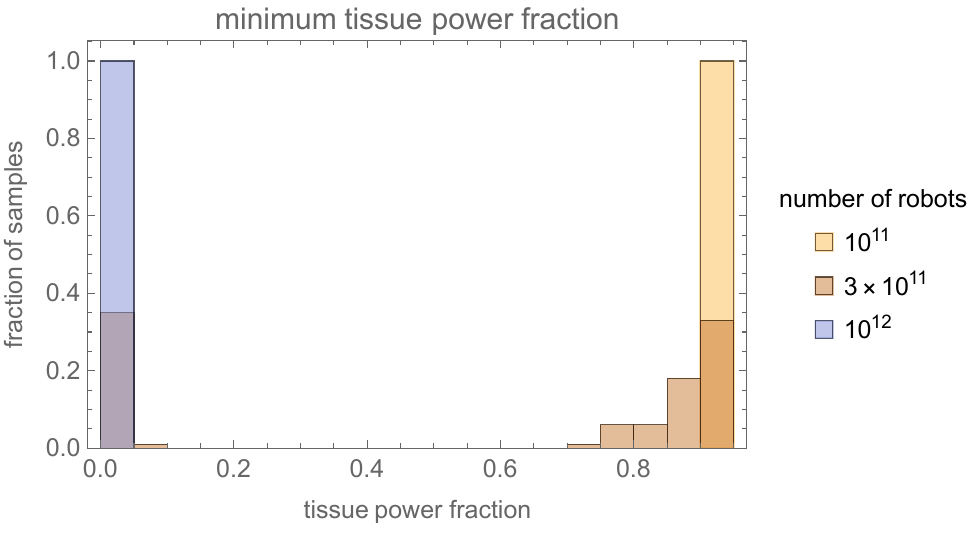}
\caption{Distribution of minimum relative tissue power in capillaries of the systemic circulation.}\figlabel{sensitivity: relative tissue power}
\end{figure}

As discussed in \sect{robot power}, the change in relative tissue power indicates how robot consumption affects tissue. In particular, the minimum value among all capillaries in the body quantifies the worst-case effect of  oxygen depletion. \fig{sensitivity: relative tissue power} shows how this minimum varies with circulation parameters. 
For $10^{11}$ or $10^{12}$ robots, the distribution is similar to that for minimum concentration at the end of segments\footnote{Capillaries mainly occur near the middle of segments (see \fig{hematocrit}), in which case concentration in a capillary is somewhat higher than at the end of segment.} shown in \fig{sensitivity: concentration}. That is, there is little variation in oxygen concentration, and hence tissue power, in capillaries in these cases. 

However, with $3\times 10^{11}$, tissue power has a bimodal distribution: for about half the parameter samples, minimum tissue power with this many robots is similar to that of $10^{11}$ robots while the other half of the samples reduce tissue power nearly as much as $10^{12}$ robots. This bimodal distribution of power arises from the nonlinear relation between oxygen concentration and tissue power~\cite{hogg21-average}, and contrasts with the wide spread of concentration values for this number of robots seen in \fig{sensitivity: concentration}.
 
As is the case for minimum concentration, as seen for the parameters of \tbl{segment parameters} in \fig{tissue power}, minimum tissue power mostly occurs in the slow spleen segment.
The significance of this reduction depends on how tolerant that tissue is of decreased oxygen and the extent those cells can obtain oxygen from the much larger flow through the fast spleen segment, e.g., due to the small vessels of these segments being close enough that tissue cells can obtain oxygen from either. This contrasts with the tissue consumption model used here which assumes each portion of tissue only receives oxygen from capillaries of a single segment.

These illustrations of the sensitivity to uncertainty in circulation parameters show that variation has little effect on average power. It also has little effect on minimum concentration or tissue power for the extremes considered here, i.e., $10^{11}$ and $10^{12}$ robots. This reinforces the conclusion from \sect{robot power} for the parameters in \tbl{segment parameters}: there is little oxygen depletion with up to $10^{11}$ robots and significant depletion with $10^{12}$ or more robots. 

When the number of robots is between these extremes, the minimum concentration and tissue power is sensitive to variation in circulation parameters. In such cases, mission plans could assume the worst case as a safety margin. Alternatively, circulation properties could be measured more precisely so as to reduce the uncertainty. 
For example, evaluating the model using patient-specific circulation parameters could determine whether that number of robots might be problematic and thus whether the mission plan will require the additional complexity of mitigation techniques discussed in \sect{mitigation}.

\section{Compensating for Circuit Variation}\sectlabel{mitigation}

The study of robots in an average circulation loop~\cite{hogg21-average} identified several ways robots could adjust their behavior to avoid extreme reductions in oxygen concentration. These methods also apply to the variable circulation loops considered here. In particular, these mitigation strategies would have the greatest benefit for robots toward the end of long circulation loops, i.e., in moderate-sized veins before they merge with veins carrying blood from shorter paths. This section describes how variation in circulation paths provides additional mitigation techniques.

\subsection{Oxygen Storage}

On-board oxygen tanks are one approach to providing power in spite of low oxygen concentration in the blood~\cite{hogg21-average}. 
This would be especially useful during long circulation loops to provide sufficient power to the robots.
Robots could reserve stored oxygen to use only on long loops where there is insufficient oxygen available in the blood.

Large tanks require that robots wait in lung capillaries long enough to fill the tanks, which significantly increases the concentration of robots in lung capillaries~\cite{hogg21-average}. 
Variation in circulation times provides another option: robots could make use of their occasional passage through short circulation loops by adding oxygen to their tanks each time they pass through a lung capillary. Robots that happen to pass through several short loops in succession would be able to completely fill their tanks before their next passage through a long loop. This procedure will not guarantee full tanks for all robots, e.g., if a robot happens to have successive passages through long loops. But on average it allows robots to take advantage of short circulation times to prepare for occasional longer loops where supplemental power from oxygen tanks is necessary.

To illustrate this technique, suppose robots collect enough oxygen to provide $100\,\picowatt$ for the $60\,\second$ average circulation time. This corresponds to storing $1.8\times 10^{10}$ oxygen molecules~\cite{hogg21-average}.
Storing this much oxygen in a tank with $10\,\nanometer$ thick walls at one-third its maximum pressure uses about one-third of the robot volume.
At the rate oxygen diffuses to the robot surface in a lung capillary, filling the tank takes about $10\,\second$~\cite{hogg21-average}.
A robot would need to pass through the lungs about 14 times to fill this tank, based on the average lung capillary transit time of $0.75\,\second$~\cite{feher17}.

However, due the Fahraeus effect discussed in \sect{hematocrit variation}, robots tend to move a bit faster than blood plasma through capillaries, giving them a bit less time to add oxygen to their tanks. 
In this model, robots require about 20 passages through the lungs to completely fill an initially empty tank, as illustrated in \fig{oxygen storage}.

\begin{figure}
\centering 
\includegraphics[width=\figwidth]{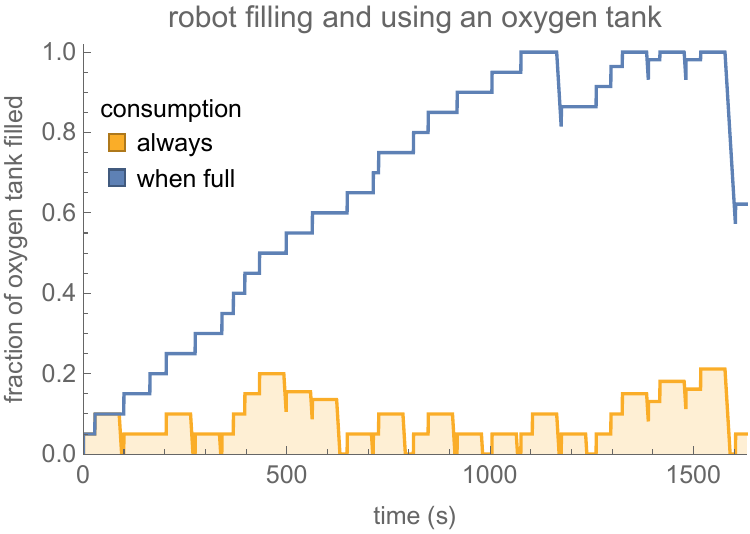}
\caption{Fraction of tank filled with oxygen vs.~time as a robot travels through randomly selected circulation paths, starting with an empty tank. The curves illustrate two methods of using oxygen in tanks, starting one minute after passing through the lungs: 1) always extract enough oxygen from the tank to provide $100\,\picowatt$, and 2) only start using the tank when it is full.}\figlabel{oxygen storage}
\end{figure}

Using stored oxygen to provide power on long circulation loops requires robots to decide when they are on a long path, and how rapidly they should consume oxygen from their tanks while on such paths. For example, suppose robots wait until one minute after their most recent passage through the lungs before starting to use their stored oxygen. This amount of time is equal to the average circulation time of all paths. Thus robots that have not returned to the lung after a minute are on a longer path than average.
Once a robot reaches this time threshold, two approaches to consuming stored oxygen are:
\begin{itemize}
\item First, robots consume oxygen to produce $100\,\picowatt$. Thus a robot on a long circulation loop gets $100\,\picowatt$, but possibly only for a part of that time if its tank is not full when starting  the loop.

\item Second, a robot only starts consuming oxygen from its tank after the tank has completely filled during multiple passes through the lung. As described above, a full tank provides $100\,\picowatt$ for $60\,\second$. In this case, robots with partially full tanks will not have supplemental power during long circulation loops.
\end{itemize}
In both cases, power production ends when either the robot reaches a lung or its oxygen tank is empty.
These methods give a trade-off of fewer robots with more power or more robots with less power in long circuits. 

Comparing these two approaches, \fig{oxygen storage} illustrates how the oxygen in a tank varies with time as a robot travels on a series of paths through the circulation over a period of about 20 minutes. The path after each lung passage is randomly selected among all paths through the network of \fig{circulation} weighted according to the blood flow in that path, as shown in \fig{paths}. This path selection corresponds to robots passively moving with the blood.
The figure shows robots alternate between brief increases in oxygen, as they pass through a lung, and decreases whenever they happen to take a path that lasts longer than a minute. For a tank requiring 20 lung transits to fill, always consuming oxygen on long paths leads to tanks storing only about $10\%$ of their capacity. On the other hand, waiting until tanks are full takes about $10\,\minute$ to fill an initially empty tank, but then maintains nearly full tanks until the robot happens to go through a particularly long circulation path.

\begin{figure}
\centering 
\includegraphics[width=\figwidth]{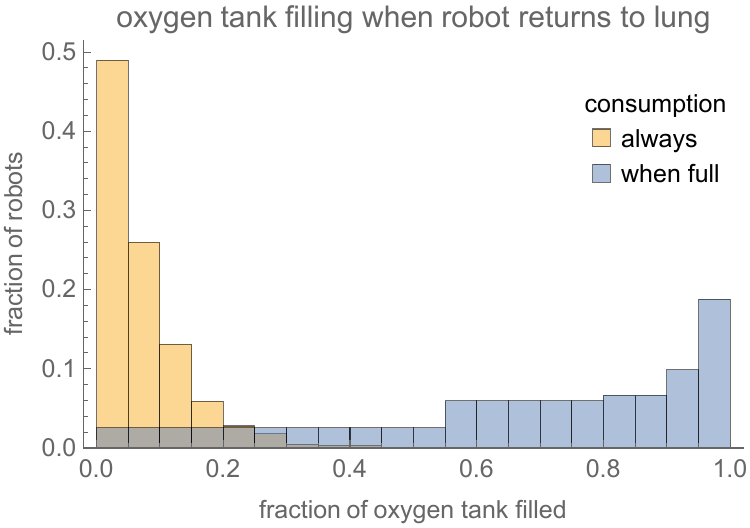}
\caption{Distribution of the fraction of tank filled with oxygen when robots return to the lung for two methods of using the oxygen from the tank starting one minute after the lung: always extract $100\,\picowatt$ from the tank, and only start using the tank when it is full.}\figlabel{oxygen storage distribution}
\end{figure}

\fig{oxygen storage} shows the behavior for robots that travel on one sequence of circulation loops. 
Different robots will move through different sequences, sometimes returning quickly to a lung and in other cases taking several minutes to return. This variation leads to a distribution in the amount of oxygen stored in the tanks. 

\fig{oxygen storage distribution} shows the resulting distribution of oxygen in tanks when robots return to a lung. 
When robots always use oxygen, their tanks are mostly less than $30\%$ full when they return to a lung. Thus, even after adding $5\%$ of tank capacity during lung capillary transit, the tank is rarely filled more than one-third of its capacity. This indicates a tank able to hold oxygen to provide $100\,\picowatt$ for $60\,\second$ that is only used during circulations longer than a minute is mostly wasted volume in robots.
On the other hand, robots that wait until their tanks are full can use most of the tank capacity. Occasionally such robots pass through circulation paths long enough to empty their tanks, and then must wait for 20 circuits to fill their tank again.

This distribution of oxygen in tanks is an example of the how a robot's history affects its capabilities. 
Exploiting this variation could allow performing missions beyond the capability of most of robots by relying on those that happen to be outliers. With large numbers of robots, even a small fraction of outliers can be many robots. 

In this example, robots determine when to use oxygen from their tanks based on the time since they leave the lung, i.e., they use a clock. This allows starting to compensate for decreased oxygen in a long circuit before the oxygen concentration gets low. Robots could instead use other criteria for when to use stored oxygen. For example, robots could attempt to maintain a specific level of power, e.g., $100\,\picowatt$, throughout their circulation: reducing their oxygen consumption when the surrounding plasma provides more than this~\cite{hogg21-average}, e.g., in arteries, and supplementing from their oxygen tanks when there is insufficient oxygen in the plasma. Or robots could save their stored oxygen until encountering specific situations that require high burst power. Such situations could include detecting a problem that requires immediate communication over long distances or if needing to move upstream in blood vessels.

An alternate use of the tank is to adjust robot power demand: supplementing with the oxygen tank when available and otherwise reducing power consumption to that available from oxygen in the plasma. For the long circuits, the relatively small fraction of robots that have returned multiple times to the lung without going on a long path, so have had an opportunity to fill their tanks much more than average, will be the robots with additional power available during the long circuits. If missions only require a portion of robots have significant power in long circulation loops, those robots that happen to start with nearly full tanks could perform those high-power tasks. For example, these high-power robots could collect information from neighboring robots and use their extra power to analyze the data and communicate a summary of those measurements to those neighbors or to an external receiver. In this way, the robots that happen to have more oxygen due to their history of short paths could provide compute or communication resources to other robots without having to create a separate class of robots for such tasks.

\subsection{Location-Specific Power Limits}

Robots can provide more oxygen to capillaries and veins by limiting their oxygen use when concentration is large (i.e., in arteries). 
Accounting for variations in the circulation allows fine-tuning this mitigation by using it only in long circulation loops where it would have significant effect. 

For example, a limit on consumption in large arteries could be relaxed in branches going to shorter circulation loops. Robots consuming more oxygen in those branches would not affect oxygen available in longer loops. This would provide, for instance, more oxygen in blood flowing to the legs while allowing robots to use more oxygen on shorter circuits where their oxygen consumption does not significantly reduce oxygen in the blood before it returns to the lungs.

Limiting power in some locations may affect mission performance, e.g., mapping the entire vasculature. However, if  the main power demands occur in specific organs or in capillaries, then limiting power during transit to those locations would not affect the mission. Thus instead of overall power limits~\cite{hogg21-average}, adjusting the limits to account for circulation variation allows robots to use higher power in a wider range of circumstances. On the other hand, restricting power only in some locations requires more complex controls for the robots. This control will benefit from some anticipation on the part of robots, depending both on the nature of a robot's task and the number of other robots.
In particular, the further upstream in the circulation a robot can determine whether it is flowing to a low-concentration region, the sooner it could limit its power. Otherwise robots may need to limit their power just in case they eventually go to those locations, even after they have passed into a branch leading to a short circulation loop where there is no need for power limits.

Robots may have some tasks that they can perform anywhere in the circulation, such as data analysis or maintenance operations. To accommodate location-specific power limits, these tasks could be relegated to short circulation loops where oxygen is plentiful. This would leave more oxygen available for location-specific tasks. 
More generally, the locations where robots have greatest effect on tissue metabolism identified in \sect{robot power} and those tissues' sensitivity to hypoxia can guide the assignment of robot tasks to different parts of the body.

\subsection{Circulation Path Selection}

Robots with locomotion and global navigation capability could pick which circulation paths they enter. For example, robots favoring shorter paths would reduce the variation in oxygen concentration that occurs when robots passively follow the blood flow. The usefulness of this approach depends on the nature of the robots' tasks, i.e., the required concentration of robots in various parts of the body. Concentrating on shorter paths would be particularly beneficial if, say, the robots are performing a therapeutic mission, such as targeted drug delivery, in an organ whose blood circulation time is relatively short. On the other hand, diagnostic or mapping missions that require robots to pass near cells of the entire body would have to accept the limited power or reduced concentration for robots in long circulation loops.
An ability to selectively avoid vessel segments may be necessary for other reasons, e.g., to avoid small slits in the spleen~\cite{freitas03}. 

Such activities require power to determine relevant locations and to take the appropriate actions.
For slits in the spleen, the robot would need to identify its location and move upstream into another vessel branch to avoid trapping in the spleen. Or, if there is significant variation in the size of the slits, a robot could expend power to search for a slit that is large enough for it to pass through.

As discussed in \sect{flow parameters}, $5\%$ of cardiac output goes to the spleen, of which $10\%$ goes through the slits. This means that, on average, a robot will circulate about $200$ times before encountering the slits, i.e., about $200\,\minute$. Missions with much shorter durations, e.g., less than an hour, will not have a significant fraction of robots reaching the slits during the mission. Longer missions require examining rarer situations leading to robot accumulation than shorter missions. Such robot accumulations could significantly deplete oxygen locally~\cite{hogg10}.

An ability for robots to select circulation paths allows a focused use of mitigation strategies. For example, only robots intended to operate in longer circulation paths would need oxygen tanks. In this case, there could be two classes of robots: those without oxygen tanks, designed for short paths, and those with tanks to supplement power during long paths. Only robots intended to operate on long paths would need to devote part of their volume to oxygen storage tanks, and only those robots would need to fill tanks while in the lungs, e.g., by sticking to capillary walls until their tanks are full or making several passes through short circuits to fill their tanks before proceeding to a long circulation path.

These two types of robots could be useful even if robots are not able to select circulation paths. In that case, robots without tanks would have little or no power in long circulation loops, and so would be temporarily unable to contribute to the mission. On the other hand,  these robots have more available internal volume for other components, such as larger computers or more sensors. These components could provide additional capabilities to robots without tanks, allowing them to contribute more to the mission while they pass through short circulation loops.

Another example of different robot types is specialized robots that can pass through spleen slits. These could be robots of smaller size, or robots that can alter their shape~\cite{freitas99} or fold~\cite{rus18} to fit through the slits in spite of their materials being too stiff to deform as red blood cells do. These robots could perform tasks within the slow compartment of the spleen, while other robots actively avoid that portion of the blood flow.

Active path selection could reduce variation in oxygen consumption in small vessels such as capillary networks. 
Small vessels will contain only a few robots at a time so will have relatively large statistical variation in their number of robots. Similar variation in the number of red cells changes how much oxygen the cells can replace.
Capillaries can have fewer red cells than average due to vessel branches selectively drawing blood from near vessel walls (plasma skimming), hence reducing the number of cells in those branches compared to other branches of similar diameter~\cite{popel05}. 
These statistical variations will only last a short time, and thereby have much less effect on nearby tissue than robots that stay for an extended time~\cite{hogg10}.
Nevertheless, the large numbers of robots and small vessels mean there will occasionally be large fluctuations in the number of robots. To avoid such cases, robots could actively reduce the variation in small vessels. Specifically, robots could preferentially select branches different from those entered by nearby downstream robots or select branches that have more than average number of red cells. This is analogous to the behavior of some immune cells in capillaries~\cite{wang20}.

\subsection{Patient Selection}

The mitigation methods described above involve changes to robot hardware, e.g., adding oxygen storage tanks, or robot behavior, e.g., selecting circulation paths. In addition, characteristics of specific patients can affect their suitability for treatment by robots using chemical power.

For instance, patients with partially occluded vessels or ineffective vein valves may have longer transit times in some parts of the body, leading to reduced oxygen available to robots and tissue. As another example, a patient whose spleen had been removed would not have the long transit through the spleen's slow compartment, thereby avoiding both that long circulation loop and the need for robots able to avoid or pass through the slits in the spleen.
Another example, applicable to the entire blood volume, is anemia, which was evaluated with the average circuit model~\cite{hogg21-average}.

In the model considered here, blood cells are fully saturated with oxygen during their passage through the lung, which is the case for healthy lung function~\cite{feher17}. However, this may not be the case for patients with compromised lung function, in which case the blood will carry less oxygen than assumed in this model. This will reduce oxygen available to robots and lead to lower oxygen concentrations when robots consume oxygen, particularly in the longer circulation routes. 
For such patients, chemical power based on oxygen from the lungs may not be adequate for the robots nor appropriate for the patient.
In this case, an alternative to relying on oxygen from the lungs is to supplement oxygen in the blood by injecting oxygen-carrying devices.
Single-use examples of such devices include bubbles enclosed in biocompatible particles of sizes from about $100\,\nanometer$ to several microns~\cite{khan18} that have been used to enhance ultrasound imaging in the body. These particles can release oxygen gradually by diffusion or rapidly in specific areas by disrupting them with resonant ultrasound.
In addition, implanted biodegradable microscopic tanks that release oxygen by diffusion have been experimentally demonstrated in tissue scaffolds~\cite{cook15}. By relying on diffusion, these tanks require no power or control, but their rate of oxygen release decreases exponentially over several hours. This approach could support robot missions of such durations, but the decreasing oxygen release could limit how well oxygen delivery matches robot demand. 
The development of more sophisticated oxygen delivery robots~\cite{freitas98} could provide greater flexibility in supplementing oxygen from the lungs.

In addition, anatomical variations among people include differences in location and connections among blood vessels. Patient-specific evaluation of robot oxygen consumption should account for any large differences in transit times due to these variations.
At smaller scales, illnesses and aging can lead to changes in the structure and function of organ microcirculation~\cite{ehling16,chen21}, which could alter the behaviors and parameters used in the model. 
Evaluating the consequences of these variations could help identify which patients will have particularly large variations in available chemical power, and adjust robot missions accordingly.

\subsection{Actively Mixing Oxygen in Merging Vessels}\sectlabel{active mixing}

The circulation model used here assumes fully mixed blood in the vessel segments. \sectA{mixing 3D} indicates that assumption may not hold at millimeter length scales for merging vessels. Thus the low concentration in one branch extends a bit into the main vessel, giving localized areas of low concentration near the walls of veins of this size, which could be detrimental to cells in those walls~\cite{hogg21-average}.

For intermediate size vessels with significantly different concentration across the vessel, robots with locomotion capability and oxygen storage tanks could increase mixing by transporting oxygen across the vessel. A simple protocol for robots to do this is to occasionally collect oxygen when they encounter relatively high concentrations, move in a series of randomly selected directions for a while, and then release the stored oxygen. This behavior would greatly increase the effective diffusion coefficient of oxygen.
Robots could enhance oxygen transport by selecting the time between direction changes based on the gradient in oxygen concentration, analogous to chemotaxis by some bacteria~\cite{berg93}. 
Alternatively, robots could move around the vessel along the wall, either in moving fluid near the wall or crawling along the vessel wall. 
While a path following the wall has a somewhat longer distance than moving across the vessel, staying near the wall avoids the higher speed flow near the center of the vessel and the complexity of moving around cells, which tend to concentrate toward the center of the vessel~\cite{freitas99}.

For faster transport, instead of randomly directed motion the robots could move directly toward parts of the vessel wall with low concentration, e.g., by comparing concentration measurements with those of other robots at different locations in the vessel if they have communication capability. Direct motion is more efficient but requires more complexity in the robot control, i.e., to determine when it is in a relatively high-concentration portion of the vessel and the direction of the low-concentration portion.
\sectA{active mixing example} gives an example of this behavior.

Even if not necessary for the safety of vein walls, oxygen transport by robots extends the accuracy of the mixing assumption used in this model to smaller length scales. This is an example of using robot behaviors to adjust local conditions so as to improve the accuracy of a model of their systemic effects on the body. Such models can help with the overall mission plan for the robots, and determine the likely effect of externally provided instructions during a treatment. Thus robots acting to improve the accuracy of the models could trade more complex individual robot activities for simpler swarm behavior design and control.

\section{Discussion}\sectlabel{discussion}

According to the model developed here, tens of billions of robots distributed throughout the blood volume do not significantly reduce oxygen available to tissues even if robots consume all oxygen reaching their surfaces. This is the case even without compensation from the body such as increased blood flow and breathing rate. This conclusion agrees with that from the average-circulation model~\cite{hogg21-average}. On the other hand, a trillion robots significantly deplete oxygen in some parts of the body, including veins returning from the legs, the liver and the slow blood transit in the spleen. 
The minimum oxygen concentration mainly occurs in moderate-sized veins toward the end of long paths prior to reaching larger veins.
These locations of low oxygen contrast with an analysis of a typical circulation loop that predicts minimum concentration in large veins and pulmonary arteries~\cite{hogg21-average}. 
This contrast shows the importance of considering variation in the circulation for evaluating minimum concentration and power for robots.

The results of the model have several consequences. One is identifying safety limits on robot oxygen consumption.
By including variation in the circulation, this model provides location-specific guidelines for how much oxygen robots can use on a sustained basis. 
If robots occasionally exceed these limits, tissue tolerance for hypoxia determines how long robots can have this larger power.

Another consequence of the variation in oxygen reduction is that robot consumption creates a chemical navigation signal, beyond those occurring normally the body~\cite{freitas99}. 
Combined with other properties, such as blood pressure, these measurements can identify the type of path and vessel (i.e., artery, capillary or vein) the robot is in.
The reduced concentration caused by robots will develop gradually after the robots start operating. This change to the robots' environment could be used as a timing signal, e.g., when to start or end tasks. 
If robots use high power in bursts and much less while they wait to see the effects of their actions or receive external commands, the oxygen concentration could repeatedly decrease and increase during these different operating regimes.

Changing oxygen concentration in space and time due to robot activity is an example of stigmergy, whereby swarms communicate indirectly through changes in their environment~\cite{bonabeau99}. Such signals could provide  guidance to robot activities. For example, lower than normal concentration in an area could indicate robots required significant power in that location because they found numerous targets requiring extensive activity. In effect, the oxygen concentration at given location is a signal that integrates robot power use in the blood upstream of that location. This is an alternative to robots actively communicating encounters with targets and other robots needing to keep track of how many such signals they receive to determine how active robots are in nearby blood vessels.

\section{Conclusion}\sectlabel{conclusion}

This model of evaluating how swarms of microscopic robots affect oxygen concentrations could be improved in several ways.

Evaluating chemical power depends on the choice of network structure (\fig{circulation}) and blood flow parameters (\tbl{segment parameters}). Refining the segments in the network and their parameters will improve the model. For example, the geometry and function of circulatory networks vary among organs, 
such as the brain~\cite{digiovanna18}, liver~\cite{haratake90} and kidney~\cite{ehling16}. Organ-specific information on vascular function and improved tools to evaluate vessel branching~\cite{tamaddon14} could improve chemical power evaluation in those organs. 
In addition, the coarse division of the circulation in \fig{circulation} combines flows with somewhat different transit times and mixing. One example is the small shunting from the systemic flow to the pulmonary vein via the bronchial circulation, which is typically 1\% of cardiac output~\cite{butler91}.
Another example is that the model assumes all robots and cells pass through capillaries to exchange oxygen with surrounding tissue. In some tissues, a small portion of the blood bypasses the capillaries~\cite{sakai13}.

This model assumes robots move with the same speed as cells. 
A direction for improving the model is identifying situations where the motion of micron-size particles differs from that of blood cells. 
One such possibility is flow through slits in the spleen, i.e., the slow spleen segment of \fig{circulation}. If robots are too large and stiff to pass through the slits, they will either need to avoid entering that part of spleen or take active measures to pass through the slits. In either case, the transit time of robots could differ from that of cells squeezing through the slits. Since this flow is a small portion of the cardiac output, this possible difference between robot and cell motion has little effect on overall power estimates from the model, but is  important for estimating power available to robots to actively adjust to the slits~\cite{freitas03}.

The consequences of robot oxygen consumption described here are based on the segment parameters in \tbl{segment parameters}.
When robots significantly alter oxygen concentrations, the body may respond by altering the speed and distribution of blood. An overall increase in blood flow applies throughout the body.  
Changes in the distribution of blood among organs is more relevant for the variation in circulation paths discussed in this paper. 
Moreover, the body's response will occur with some delay after robots deplete oxygen. If robot power use varies significantly during a mission, oxygen concentration will alternate between higher and lower values. In that case, the delay in response could lead to instability in respiration such as seen with Cheyne-Stokes breathing~\cite{francis00}.
Evaluating these changes will be especially important at the larger end of the range of robot numbers considered here, due to their significant depletion of oxygen in some parts of the body, and for applications requiring robot activity over extended periods of time.

This paper shows how large-scale variation in blood circulation time and hematocrit profiles affect the consequences of robot oxygen consumption. This analysis is particularly important for extreme properties such as minimum power available to robots and minimum oxygen concentration anywhere in the body. This study also provides some approaches to modifying robot designs and tasks to compensate for this variation.
This evaluation can help identify the suitability of chemical power for swarms of robots operating in various locations in the body.

\section*{Acknowledgements}

I have benefited from discussions with Robert Freitas, Ralph Merkle, James Ryley and Jeff Semprebon.

\newpage
\appendix

\numberwithin{equation}{section}
\numberwithin{figure}{section}
\numberwithin{table}{section}

\begin{center}
{\huge \textbf{Appendix}}
\end{center}

\section{Mixing in Merging Vessels}\sectlabel{mixing 3D}

The network circulation model assumes oxygen is fully mixed throughout each segment cross section.
This assumption allows treating each segment as a one-dimensional vessel rather than explicitly modeling variation in oxygen concentration over the segment's cross section. 
This appendix evaluates the applicability of this assumption using a more realistic three-dimensional model of merging flows.

\subsection{Merging of Millimeter-Sized Vessels}

\begin{figure}[h]
\centering 
\includegraphics[width=\figwidth]{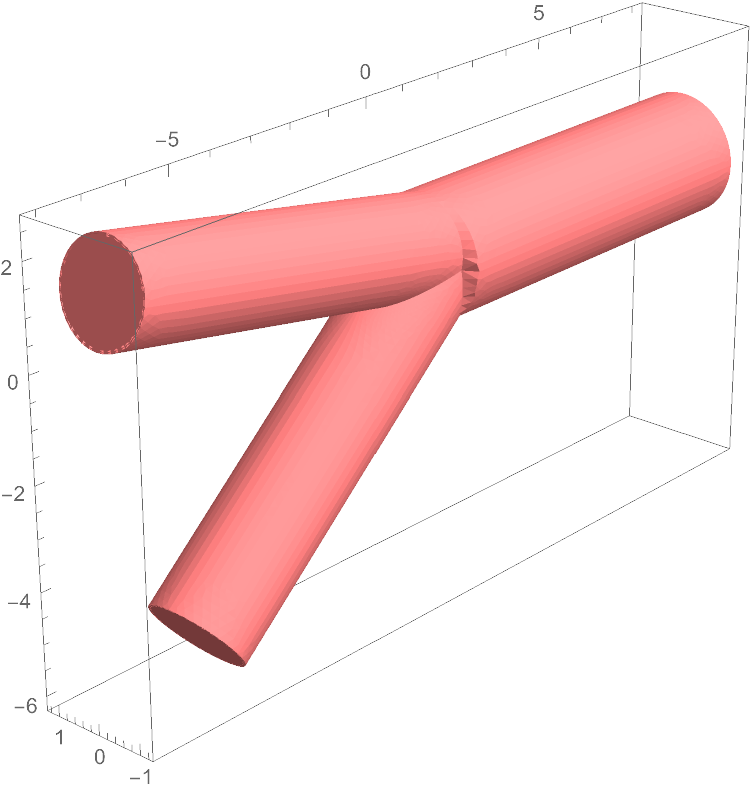}  
\caption{Example of merging vessels: the branches, each with diameter of $2\,\millimeter$, merge into a vessel with diameter of $2.5\,\millimeter$.
Numbers along the axes are positions in millimeters measured from an origin at the center of the merging vessels.}\figlabel{merging vessels}
\end{figure}

\fig{merging vessels} is an example of merging vessels. 
As is typical of merging blood vessels, the total cross section of the two branches exceeds that of the main vessel, so flow speed in the branches is somewhat slower than in the main vessel. Flow in vessels of this size is laminar. As boundary conditions on the flow for this example, we specify Poiseuille flow exiting the main vessel with average speed $2.5\,\millimeter/\second$ and the inlet pressure at both branches is the same. 
\fig{merging flow} shows how the flow from the branches combines into the main vessel. Due to the laminar nature of the flow, there is little mixing of fluid from the two branches. These vessels correspond to the merging of small veins, so there is no oxygen consumption by tissue.

\begin{figure}
\centering 
\includegraphics[width=\figwidth]{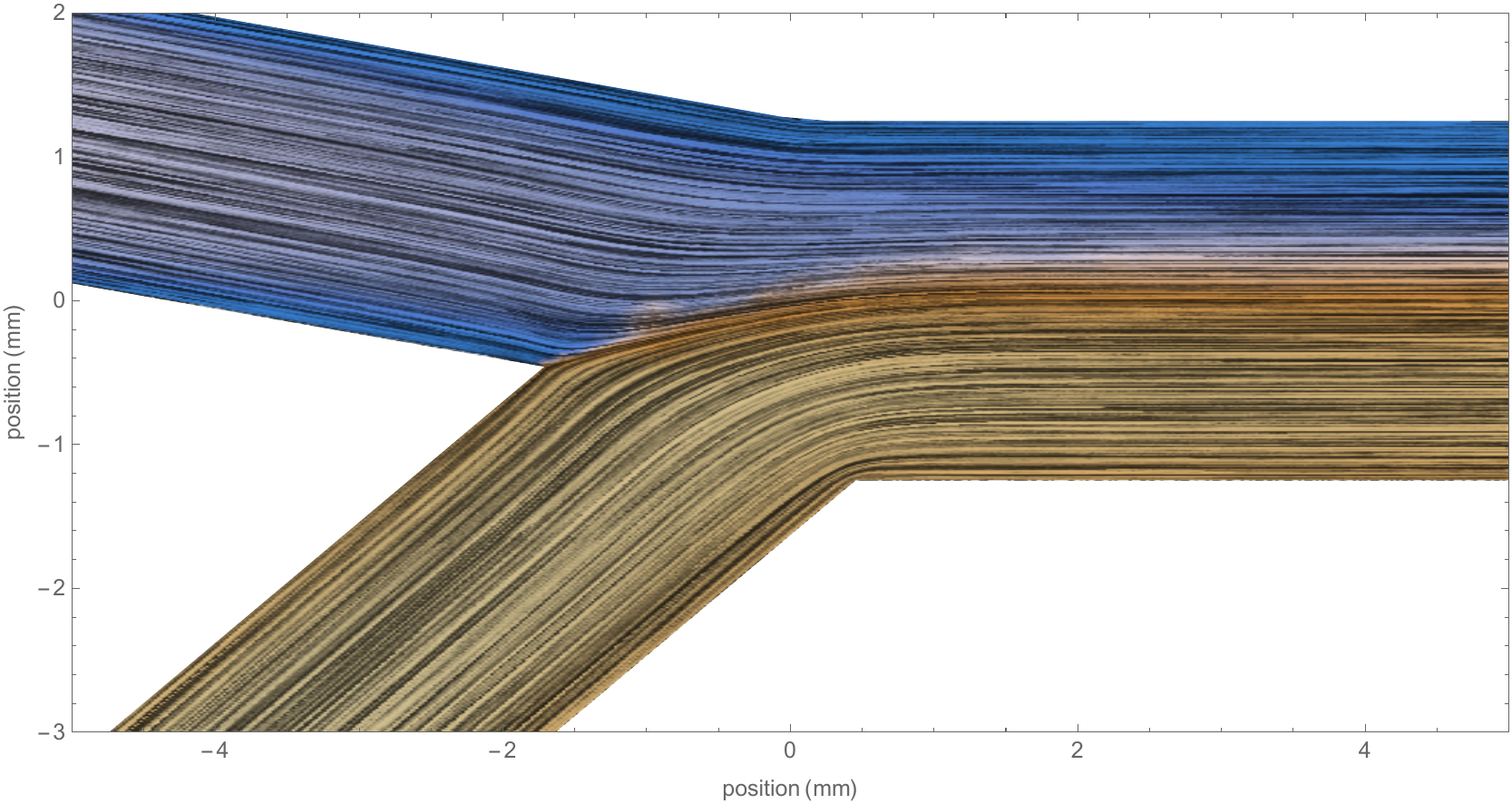} 
\caption{Fluid flow on the symmetry plane cross section of the vessels close to where the vessels merge. Numbers along the axes are positions in millimeters.}\figlabel{merging flow}
\end{figure}

To evaluate oxygen mixing, consider two incoming flows with substantially different low oxygen concentrations: $0.5$ and $2.0$ times $10^{22}\,\molecule/\meter^3$, respectively, and $10^{12}$ robots in the blood. 
These different concentrations could arise, for example, from blood returning to veins from circulation paths of significantly different lengths.

Oxygen in the fluid moves by convection with the fluid's motion as well as diffusion. For the sizes and flow speeds considered here, diffusion is a minor contribution to changing concentration. In particular, the Peclet number characterizes the relative importance of convection and diffusion~\cite{squires05}:
\begin{equation}\eqlabel{Peclet}
\Pec = \frac{v d}{ \Doxygen}
\end{equation}
where $v$ is the flow speed, $d$ a characteristic distance and $\Doxygen$ the diffusion coefficient. For flow through a vessel of diameter $d$, \Pec\ is the approximate number of vessel diameters required for diffusion to spread across the vessel. For motion along the vessel, the distance at which $\Pec \approx 1$, i.e., $d = \Doxygen/v$, is the distance at which diffusion and convection have about the same effect on mass transport in a moving fluid. At significantly longer distances, convection is the dominant effect.

In this case, the mixing due to the motion of blood cells is a minor addition to the diffusion of small molecules, such as oxygen, in the blood~\cite{freitas99}.
Thus \eq{Peclet} gives $\Pec = 3000$, so oxygen mainly flows along the vessel with relatively little diffusion across it.
This behavior leads to considerable variation in oxygen concentration across the vessel: robots in slow moving blood near the walls deplete oxygen much more than robots near the center of the vessel, where faster flow replaces the blood before robots have drawn down the oxygen as much as robots near the vessel wall.

\begin{figure}
\centering 
\includegraphics[width=\figwidth]{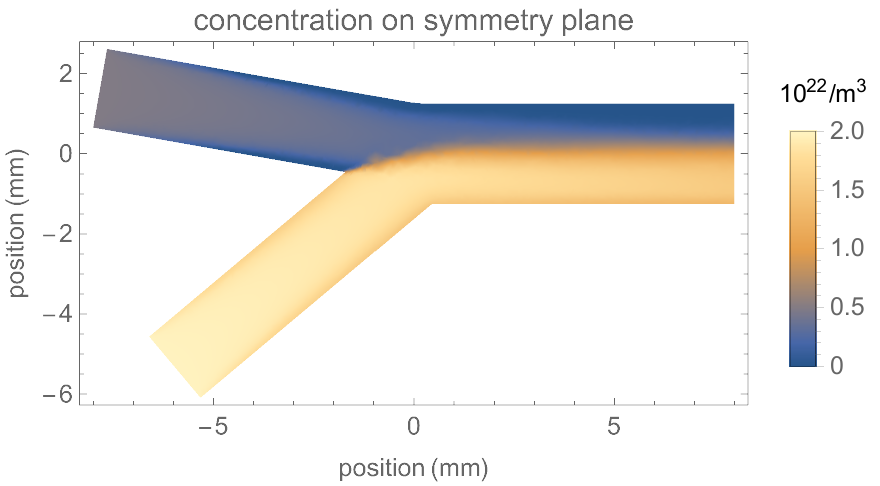} 
\caption{Oxygen concentration on a cross section through the vessels with diameter of about $2\,\millimeter$ on the symmetry plane with $10^{12}$ robots in the bloodstream, each consuming all oxygen reaching its surface.}\figlabel{merging vessels concentration}
\end{figure}

\begin{figure}
\centering 
\includegraphics[width=\figwidth]{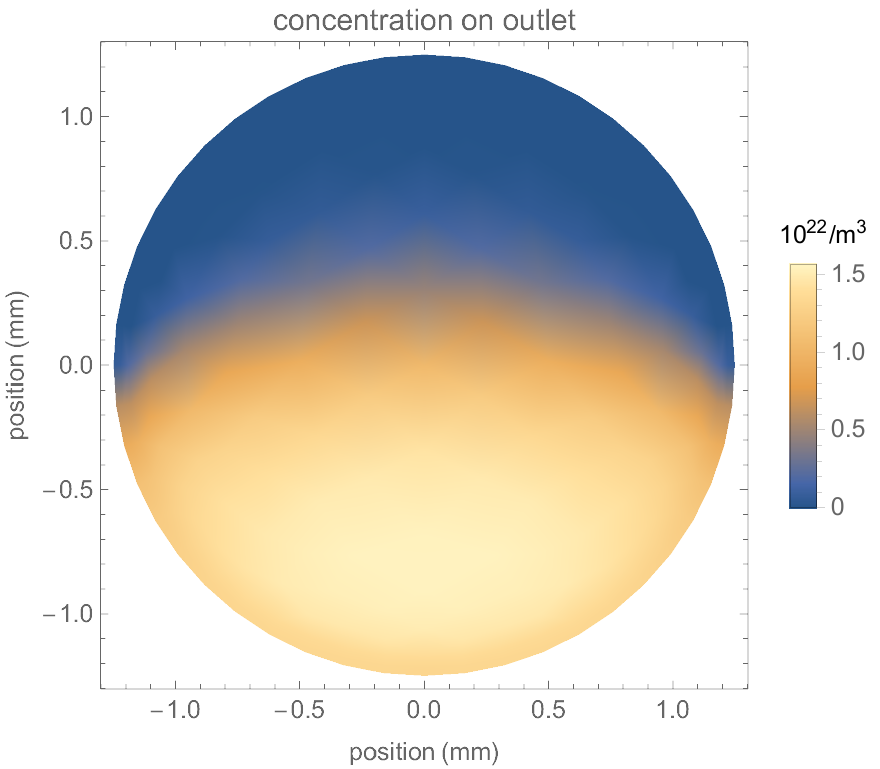} 
\caption{Oxygen concentration at the outlet with $10^{12}$ robots in the bloodstream, each consuming all oxygen reaching its surface.}\figlabel{merging vessels outlet concentration}
\end{figure}

\fig{merging vessels concentration} shows the resulting concentration on the symmetry plane of the merging vessels, and \fig{merging vessels outlet concentration} shows the concentration on the outlet cross section. These figures show that oxygen is not fully mixed across the vessel a few millimeters downstream of the merge. Thus the assumption of fully mixed flow does not hold over the millimeter size scales of this example. Instead, as indicated by the Peclet number, mixing throughout the vessel requires a few centimeters. 

This example is for merging of moderate size veins. Larger vessels can have more complex laminar flows~\cite{freitas99}, which could lead to more rapid mixing than in this example.

As illustrated by this example, mixing in millimeter-sized vessels is mainly due to flow, not diffusion. Thus, these intermediate-size vessels have smooth flow that does not mix much. 
For investigating major variations in circulation paths, these observations motivate focusing on segments that aggregate vessels over lengths of centimeters or more, as used in \fig{circulation}.

\subsection{Merging of Sub-Millimeter Vessels}

\begin{figure}
\centering 
\includegraphics[width=\figwidth]{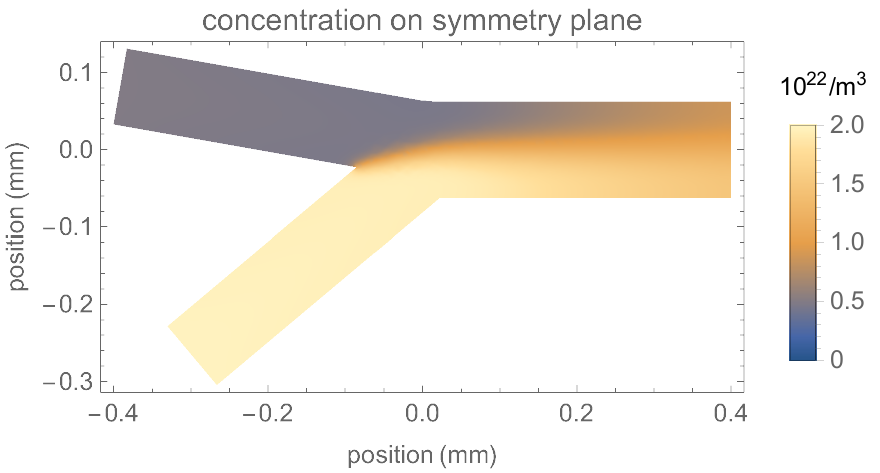} 
\caption{Oxygen concentration on a cross section through vessels with diameter of about $100\,\micron$ on the symmetry plane with $10^{12}$ robots in the bloodstream, each consuming all oxygen reaching its surface.}\figlabel{merging vessels concentration smaller}
\end{figure}

Over small distances, diffusion is relatively fast. This leads to more rapid mixing in smaller vessels than the example in \fig{merging vessels}. To illustrate this behavior, consider merging vessels with the same geometry as \fig{merging vessels} but $5\%$ of the size and $30\%$ the flow speed.
In this case, the Peclet number is 45 leading to nearly complete mixing shortly after the merge, as shown in \fig{merging vessels concentration smaller}.

The fast oxygen diffusion on this scale contrasts with slower diffusion of cells and robots in the blood~\cite{freitas99}.
Thus the rapid mixing of oxygen through the blood plasma could increase oxygen in parts of the flow with depleted cells arriving from the low-concentration vessel (i.e., the vessel at the upper left of \fig{merging vessels concentration smaller}). In this case, the cells will absorb some of that oxygen while cells in the high-concentration portion of the vessel release more oxygen, thereby somewhat delaying the mixing. However, hemoglobin binding kinetics is fast compared to these flows~\cite{clark85}, so this is a minor effect.

\subsection{Actively Increasing Mixing}\sectlabel{active mixing example}

\fig{merging vessels concentration} is an example of merging vessels with different oxygen concentrations. For vessels of this size and flow speed, oxygen does not mix throughout the vessel over the millimeter distances shown in the figure. This could lead to low concentration on part of the vessel wall. To avoid such localized regions of low concentration, \sect{active mixing} described how robots could actively increase mixing in vessels by transporting oxygen across them.

For the example of \fig{merging vessels concentration}, in the high-concentration part of the vessel, $5\times 10^8$ oxygen molecules reach a robot's surface each second~\cite{hogg21-average}.
A robot could collect molecules at this rate for one second and store them in a tank using about $1\%$ of the robot's volume~\cite{freitas99}.
The robot could carry this oxygen across the vessel, about a distance of $2.5\,\millimeter$ in this example, at, say, a speed of $1\,\millimeter/\second$.

This motion requires power to overcome drag on the robot due to fluid viscosity~\cite{purcell77}. For example, the drag on a sphere of radius $r$ moving at speed $v$ in fluid with viscosity $\eta$ is $6 \pi \eta a v$~\cite{berg93}. Locomotion in viscous fluids typically has efficiency of a few percent~\cite{freitas99,hogg14}. For a conservative estimate, assume fuel cells are $50\%$ efficient and locomotion is only $1\%$ efficient. For this example, these efficiencies mean robots require a few picowatts while they move across the vessel. Robots could obtain this energy by using a few percent of the oxygen molecules stored in their tanks. 

Suppose $n$ robots are uniformly distributed in the blood volume. Then a small volume $\Delta V$ near the high-concentration side of the vessel contains $n \Delta V/\bloodVolume$ robots. With each robot carrying $m=5\times 10^8$ oxygen molecules across the vessel in time $t=1\,\second$, these robots increase oxygen concentration on the other side of the vessel at a rate
\begin{equation}
\frac{dc}{dt} = \frac{n}{ \bloodVolume} \frac{m}{t}
\end{equation}
With $n=10^{12}$ robots in the blood, $dc/dt = 4 \times 10^{22}/\meter^3/\second$. Thus the robots mix concentration across the vessel in \fig{merging vessels concentration} in less than a second.

Instead of direct transport, robots could use a simpler protocol of randomly directed motion to move oxygen in the vessels. The robots would, in effect, increase the oxygen diffusion coefficient. For instance, robots moving at $v=1\,\millimeter/\second$ and randomly changing directions at random times that are exponentially distributed with mean $\tau=0.1\,\second$, have diffusion coefficient $D=v^2 \tau/3$~\cite{berg93} which, in this case, is about 20 times larger than oxygen diffusion in blood plasma.

During its transit across the vessel, the robot also moves downstream with the fluid flow by a few millimeters (unless it uses additional power to move upstream against the flow). 
This downstream motion means oxygen from the high-concentration side of the vessel increases concentration at the other side a few millimeters downstream of where robots collect oxygen. This is a much shorter mixing distance than shown in \fig{merging vessels concentration}. Alternatively, robots could stay close to the vessel wall as they move from one side to the other, thereby avoiding the faster flow near the center of the vessel and not moving downstream as far.

This example shows that active oxygen transport by robots in millimeter-size vessels can increase mixing to an extent similar to that seen in smaller vessels in \fig{merging vessels concentration smaller}.

\section{Segments and Merging Vessels}\sectlabel{segments and vessels}

\begin{figure}
\centering 
\begin{tabular}{cc}
\includegraphics[width=\mfigwidth]{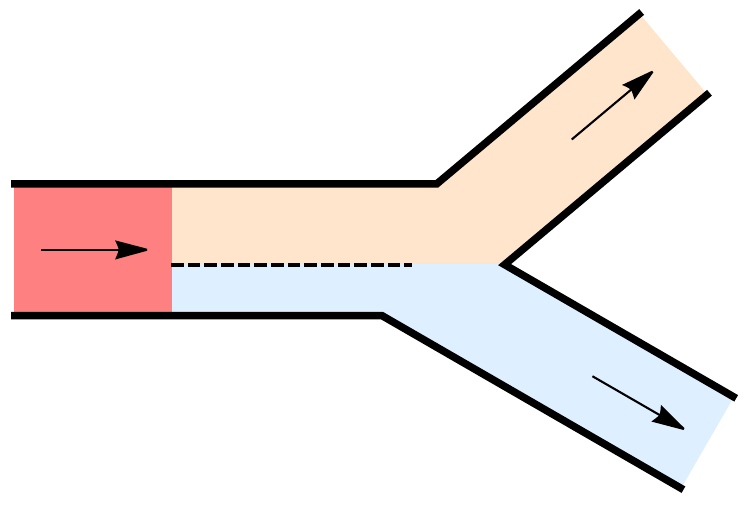}	&	\includegraphics[width=\mfigwidth]{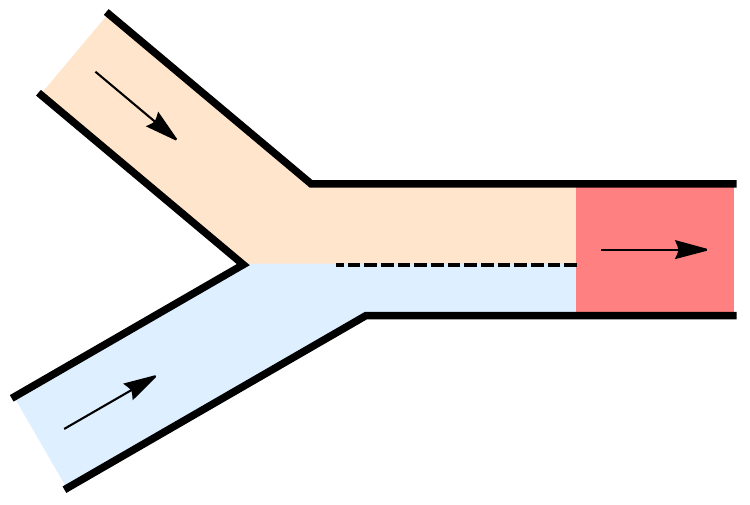} \\
(a) & (b) \\
\end{tabular}
\caption{Choice of segment extent near splitting and merging vessels. The colors indicate parts of three segments joining near the vessel branch. The dashed line indicates an extension of the segments into the main vessel. Arrows indicate direction of blood flow. (a) A vessel splits into two branches. (b) Two vessels merge into one.}\figlabel{segment joins}
\end{figure}

The network model of \fig{circulation} consists of segments describing flows that branch and merge throughout the body. A direct interpretation of these segments is that they correspond to branching of major blood vessels. However, for simplicity and to address the mixing issue described in \sectA{mixing 3D}, segments can include a portion of the flow in larger vessels, so that segments do not necessarily correspond directly to blood vessel branches. Thus, the names associated with the segments in \fig{circulation} are only rough descriptions of the portion of the flow carried by the segment.

For instance, mixing blood from hepatic artery and portal vein occurs in liver capillaries~\cite{feher17}. Thus the hepatic and portal segments include the branching vessels to the liver capillaries, and so include portions of the liver. Since a large portion of vascular resistance occurs in arterioles, this means that a significant portion of the vascular resistance of the liver is, in this model, included in the hepatic and portal segments.

Moreover, when a large vessel splits, the concentration in the branches is the same as in the main vessel (see \eq{branch split}). Thus, for evaluating how robots affect concentration, a segment including one of the branches could include any convenient portion of the blood in the larger upstream vessels. Schematically, \fig{segment joins}a shows how segments can extend upstream into larger vessels, by including the portion of the blood in the larger vessel that eventually flows into a smaller vessel within the segment.
This flexibility in defining segments allows simplifying the network model of \fig{circulation} by associating portions of the flow in large arteries with segments containing flow in smaller vessels, rather than treating large arteries, e.g., the aorta, as separate segments.

As an example, the hepatic artery has about $5\%$ of the cardiac output (see \tbl{segment parameters}). The hepatic segment of \fig{circulation} contains blood flow starting from when it leaves the heart. Thus the transit time through the hepatic segment includes the time taken to pass through the larger arteries that branch to the hepatic artery. 
The flow in these large vessels is faster than in the hepatic artery, so this extension contributes proportionally less to transit time than it does to length and blood volume.

In merging vessels, \sectA{mixing 3D} indicates that, in some cases, oxygen does not fully mix until somewhat downstream of the merge. 
The portion of flow in merged vessels that has not mixed acts as if the flow is still in distinct vessels. In effect, the distance in the merged vessels required to achieve mixing (as assumed for the one-dimensional aggregate flow of this model, and in particular \eq{branch merge}) means the effective length of the merging branches is a bit longer than their physical length, and the length of the merged vessel is somewhat smaller. This variation is included in the model by defining the merging segments to extend far enough downstream of the merge to reach a location where oxygen has sufficiently mixed, as indicated schematically in \fig{segment joins}b. The model incorporates such adjustment by increasing the transit time and vascular resistance associated with the segment in \tbl{segment parameters}.



\clearpage

\begin{thebibliography}{10}

\bibitem{ahlborg91}
G.~Ahlborg and M.~Jensen-Urstad.
\newblock Arm blood flow at rest and during arm exercise.
\newblock {\em J. of Applied Physiology}, 70:928--933, 1991.

\bibitem{amar15}
Achraf~Ben Amar, Ammar~B. Kouki, and Hung Cao.
\newblock Power approaches for implantable medical devices.
\newblock {\em Sensors}, 15:28889--28914, 2015.

\bibitem{an11}
Liang An et~al.
\newblock Alkaline direct oxidation fuel cell with non-platinum catalysts
  capable of converting glucose to electricity at high power output.
\newblock {\em J. of Power Sources}, 196:186--190, 2011.

\bibitem{angleys20}
Hugo Angleys and Leif Ostergaard.
\newblock {Krogh's} capillary recruitment hypothesis, 100 years on: Is the
  opening of previously closed capillaries necessary to ensure muscle
  oxygenation during exercise?
\newblock {\em American J. of Physiology}, 318:H425--H447, 2020.

\bibitem{bazaka13}
Kateryna Bazaka and Mohan~V. Jacob.
\newblock Implantable devices: Issues and challenges.
\newblock {\em Electronics}, 2:1--34, 2013.

\bibitem{berg93}
Howard~C. Berg.
\newblock {\em Random Walks in Biology}.
\newblock Princeton Univ. Press, 2nd edition, 1993.

\bibitem{bonabeau99}
Eric Bonabeau, Marco Dorigo, and Guy Theraulaz.
\newblock {\em Swarm Intelligence: From Natural to Artificial Systems}.
\newblock Oxford University Press, Oxford, 1999.

\bibitem{brooks20}
Allan~M. Brooks and Michael~S. Strano.
\newblock A conceptual advance that gives microrobots legs.
\newblock {\em Nature}, 584:530--531, 2020.

\bibitem{butler91}
John Butler.
\newblock The bronchial circulation.
\newblock {\em Physiology}, 6:21--25, 1991.

\bibitem{carreau11}
Aude Carreau et~al.
\newblock Why is the partial oxygen pressure of human tissues a crucial
  parameter? {S}mall molecules and hypoxia.
\newblock {\em J. of Cellular and Molecular Medicine}, 15:1239--1253, 2011.

\bibitem{cataldi17}
Mauro Cataldi et~al.
\newblock Emerging role of the spleen in the pharmacokinetics of monoclonal
  antibodies, nanoparticles and exosomes.
\newblock {\em Intl. J of Molecular Sciences}, 18:1249, 2017.

\bibitem{chaudhuri03}
Swades~K. Chaudhuri and Derek~R. Lovley.
\newblock Electricity generation by direct oxidation of glucose in mediatorless
  microbial fuel cells.
\newblock {\em Nature Biotechnology}, 21:1229--1232, 2003.

\bibitem{chen21}
Junyu Chen et~al.
\newblock High-resolution {3D} imaging uncovers organ-specific vascular control
  of tissue aging.
\newblock {\em Science Advances}, 7:eabd7819, 2021.

\bibitem{clark85}
Alfred Clark, Jr. et~al.
\newblock Oxygen delivery from red cells.
\newblock {\em Biophysical Journal}, 47:171--181, 1985.

\bibitem{cook15}
Colin~A. Cook et~al.
\newblock Oxygen delivery from hyperbarically loaded microtanks extends cell
  viability in anoxic environments.
\newblock {\em Biomaterials}, 52:376--384, 2015.

\bibitem{deussen12}
Andreas Deussen et~al.
\newblock Mechanisms of metabolic coronary flow regulation.
\newblock {\em J. of Molecular and Cellular Cardiology}, 52:794--801, 2012.

\bibitem{digiovanna18}
A.~P. {Di Giovanna} et~al.
\newblock Whole-brain vasculature reconstruction at the single capillary level.
\newblock {\em Scientific Reports}, 8:12573, 2018.

\bibitem{dock61}
Donald~S. Dock et~al.
\newblock The pulmonary blood volume in man.
\newblock {\em J. of Clinical Investigation}, 40:317--328, 1961.

\bibitem{dong07}
Lixin Dong and Bradley~J. Nelson.
\newblock Robotics in the small. part {II}: Nanorobotics.
\newblock {\em IEEE Robotics \& Automation Magazine}, 14:111--121, 2007.

\bibitem{dreyfus05}
Remi Dreyfus et~al.
\newblock Microscopic artificial swimmers.
\newblock {\em Nature}, 437:862--865, 2005.

\bibitem{effros67}
Richard~M. Effros et~al.
\newblock Vascular and extravascular volumes of the kidney of man.
\newblock {\em Circulation Research}, 20:162--173, 1967.

\bibitem{ehling16}
J.~Ehling et~al.
\newblock Quantitative micro-computed tomography imaging of vascular
  dysfunction in progressive kidney diseases.
\newblock {\em J. of the American Society of Nephrology}, 27:520--532, 2016.

\bibitem{feher17}
Joseph Feher.
\newblock {\em Quantitative Human Physiology}.
\newblock Academic Press, 2nd edition, 2017.

\bibitem{francis00}
Darrel~P. Francis et~al.
\newblock Quantitative general theory for periodic breathing in chronic heart
  failure and its clinical implications.
\newblock {\em Circulation}, 102:2214--2221, 2000.

\bibitem{freitas98}
Robert~A. {Freitas Jr.}
\newblock Exploratory design in medical nanotechnology: A mechanical artificial
  red cell.
\newblock {\em Artificial Cells, Blood Substitutes and Immobilization
  Biotechnology}, 26:411--430, 1998.

\bibitem{freitas99}
Robert~A. {Freitas Jr.}
\newblock {\em Nanomedicine}, volume {I}: Basic Capabilities.
\newblock Landes Bioscience, Georgetown, TX, 1999.
\newblock Available at www.nanomedicine.com/NMI.htm.

\bibitem{freitas03}
Robert~A. {Freitas Jr.}
\newblock {\em Nanomedicine}, volume {IIA}: Biocompatibility.
\newblock Landes Bioscience, Georgetown, TX, 2003.
\newblock Available at www.nanomedicine.com/NMIIA.htm.

\bibitem{galton1883}
Francis Galton.
\newblock {\em Inquiries into Human Faculty, and Its Development}.
\newblock Macmillan, London, 1883.

\bibitem{gibson46}
John~G. {Gibson 2nd} et~al.
\newblock The distribution of red cells and plasma in large and minute vessels
  of the normal dog, determined by radioactive isotopes of iron and iodine.
\newblock {\em The J. of Clinical Investigation}, 25:848--857, 1946.

\bibitem{gogova10}
Zuzand Gogova, Jiri Hanika, and Jozef Markos.
\newblock Optimal design of a multifunctional reactor for catalytic oxidation
  of glucose with fast catalyst deactivation.
\newblock In Alisson~V. Brito, editor, {\em Dynamic Modelling}, chapter~12,
  pages 209--232. InTech, Rijeka, Croatia, 2010.

\bibitem{groom02}
Alan~C. Groom, Ian~C. MacDonald, and Eric~E. Schmidt.
\newblock Splenic microcirculatory blood flow and function with respect to red
  blood cells.
\newblock In A.~J. Bowdler, editor, {\em The Complete Spleen}, pages 23--50.
  Humana Press, Totowa, NJ, 2002.

\bibitem{happel83}
John Happel and Howard Brenner.
\newblock {\em Low {Reynolds} Number Hydrodynamics}.
\newblock Kluwer, The Hague, 2nd edition, 1983.

\bibitem{haratake90}
J.~Haratake et~al.
\newblock Scanning electron microscopic examinations of microvascular casts of
  the rat liver and bile duct.
\newblock {\em J. of UOEH}, 12:19--28, 1990.

\bibitem{hashemi20}
A.~{Hashemi Talkhooncheh} et~al.
\newblock A fully-integrated biofuel-cell-based energy harvester with 86\% peak
  efficiency and {0.25V} minimum input voltage using source-adaptive {MPPT}.
\newblock In {\em Proc. of the 2020 IEEE Custom Integrated Circuits Conference
  (CICC)}, pages 1--4. IEEE, 2020.

\bibitem{hogg14}
Tad Hogg.
\newblock Using surface-motions for locomotion of microscopic robots in viscous
  fluids.
\newblock {\em J. of Micro-Bio Robotics}, 9:61--77, 2014.

\bibitem{hogg21-average}
Tad Hogg.
\newblock Chemical power for swarms of microscopic robots in the bloodstream.
\newblock Technical report, arxiv.org/abs/2301.11286, 2023.

\bibitem{hogg10}
Tad Hogg and Robert~A. {Freitas Jr.}
\newblock Chemical power for microscopic robots in capillaries.
\newblock {\em Nanomedicine: Nanotechnology, Biology, and Medicine},
  6:298--317, 2010.

\bibitem{holland98}
Christy~K. Holland et~al.
\newblock Lower extremity volumetric arterial blood flow in normal subjects.
\newblock {\em Ultrasound in Medicine and Biology}, 24:1079--1086, 1998.

\bibitem{huang96}
W.~Huang et~al.
\newblock Morphometry of the human pulmonary vasculature.
\newblock {\em J. of Applied Physiology}, 81:2123--2133, 1996.

\bibitem{khan18}
Muhammad~Saad Khan et~al.
\newblock Oxygen-carrying micro/nanobubbles: Composition, synthesis techniques
  and potential prospects in photo-triggered theranostics.
\newblock {\em Molecules}, 23:2210, 2018.

\bibitem{krogh19}
August Krogh.
\newblock The number and distribution of capillaries in muscles with
  calculations of the oxygen pressure head necessary for supplying the tissue.
\newblock {\em J. of Physiology}, 52:409--415, 1919.

\bibitem{krogh22}
August Krogh.
\newblock {\em The Anatomy and Physiology of Capillaries}.
\newblock Yale University Press, New Haven, CT, 1922.

\bibitem{lautt09}
W.~Wayne Lautt.
\newblock {\em Hepatic Circulation: Physiology and Pathophysiology}.
\newblock Morgan \& Claypool Life Sciences, San Rafael, CA, 2009.

\bibitem{linderkamp80}
Otwin Linderkamp et~al.
\newblock Blood volume and hematocrit in various organs in newborn piglets.
\newblock {\em Pediatric Research}, 14:1324--1327, 1980.

\bibitem{macdonald91}
I.~C. MacDonald, E.~E. Schmidt, and A.~C. Groom.
\newblock The high splenic hematocrit: A rheological consequence of red cell
  flow through the reticular meshwork.
\newblock {\em Microvascular Research}, 42:60--76, 1991.

\bibitem{martel07}
Sylvain Martel.
\newblock The coming invasion of the medical nanorobots.
\newblock {\em Nanotechnology Perceptions}, 3:165--173, 2007.

\bibitem{mcHedlishvili87}
George McHedlishvili and Manana Varazashvili.
\newblock Hematocrit in cerebral capillaries and veins under control and
  ischemic conditions.
\newblock {\em J. of Cerebral Blood Flow \& Metabolism}, 7:739--744, 1987.

\bibitem{mizuno03}
Masaki Mizuno et~al.
\newblock Regional differences in blood volume and blood transit time in
  resting skeletal muscle.
\newblock {\em Japanese J. of Physiology}, 53:467--470, 2003.

\bibitem{morris01}
Kelly Morris.
\newblock Macrodoctor, come meet the nanodoctors.
\newblock {\em The Lancet}, 357:778, March 10 2001.

\bibitem{morris57}
Leslie~E. Morris and Herrman~L. Blumgart.
\newblock Velocity of blood flow in health and disease.
\newblock {\em Circulation}, XV:448--460, 1957.

\bibitem{nelson10}
Bradley~J. Nelson, Ioannis~K. Kaliakatsos, and Jake~J. Abbott.
\newblock Microrobots for minimally invasive medicine.
\newblock {\em Annual Review of Biomedical Engineering}, 12:55--85, 2010.

\bibitem{nylin47}
Gustav Nylin.
\newblock Circulatory blood volume of some organs.
\newblock {\em American Heart Journal}, 34:174--179, 1947.

\bibitem{peters18}
A.~Michael Peters.
\newblock The precise physiological definition of tissue perfusion and
  clearance measured from imaging.
\newblock {\em European Journal of Nuclear Medicine and Molecular Imaging},
  45:1139--1141, 2018.

\bibitem{poole13}
D.~C. Poole et~al.
\newblock Skeletal muscle capillary function: contemporary observations and
  novel hypotheses.
\newblock {\em Experimental Physiology}, 98:1645--1658, 2013.

\bibitem{popel89}
Aleksander~S. Popel.
\newblock Theory of oxygen transport to tissue.
\newblock {\em Critical Reviews in Biomedical Engineering}, 17:257--321, 1989.

\bibitem{popel05}
Aleksander~S. Popel and Paul~C. Johnson.
\newblock Microcirculation and hemorheology.
\newblock {\em Annual Review of Fluid Mechanics}, 37:43--69, 2005.

\bibitem{purcell77}
Edward~M. Purcell.
\newblock Life at low {Reynolds} number.
\newblock {\em American Journal of Physics}, 45:3--11, 1977.

\bibitem{rajendran24}
Shishir Rajendran et~al.
\newblock Nanorobotics in medicine: A systematic review of advances,
  challenges, and future prospects with a focus on cell therapy, invasive
  surgery, and drug delivery.
\newblock {\em Precision Nanomedicine}, 7:1221--1232, 2024.

\bibitem{rapoport12}
Benjamin~I. Rapoport, Jakub~T. Kedzierski, and Rahul Sarpeshkar.
\newblock A glucose fuel cell for implantable brain-machine interfaces.
\newblock {\em PLoS ONE}, 7(6):e38436, 2012.

\bibitem{rus18}
Daniela Rus and Michael~T. Tolley.
\newblock Design, fabrication and control of origami robots.
\newblock {\em Nature Reviews Materials}, 3:101--112, 2018.

\bibitem{sakai13}
Tatsuo Sakai and Yasue Hosoyamada.
\newblock Are the precapillary sphincters and metarterioles universal
  components of the microcirculation? an historical review.
\newblock {\em J. of Physiological Sciences}, 63:319--331, 2013.

\bibitem{savage09}
Sam~L. Savage.
\newblock {\em The Flaw of Averages}.
\newblock John Wiley, 2009.

\bibitem{sourbron09}
Steven Sourbron et~al.
\newblock Quantification of cerebral blood flow, cerebral blood volume, and
  blood-brain-barrier leakage with {DCE-MRI}.
\newblock {\em Magnetic Resonance in Medicine}, 62:205--217, 2009.

\bibitem{squires05}
Todd~M. Squires and Stephen~R. Quake.
\newblock Microfluidics: Fluid physics at the nanoliter scale.
\newblock {\em Reviews of Modern Physics}, 77:977--1026, 2005.

\bibitem{tamaddon14}
H.~Tamaddon et~al.
\newblock A new approach to blood flow simulation in vascular networks.
\newblock {\em Computer Methods in Biomechanics and Biomedical Engineering},
  19:673--685, 2014.

\bibitem{wadenvik88}
Hans Wadenvik and Jack Kutti.
\newblock The spleen and pooling of blood cells.
\newblock {\em European J of Haematology}, 41:1--5, 1988.

\bibitem{wang20}
Xiao Wang et~al.
\newblock Chemotaxing neutrophils enter alternate branches at capillary
  bifurcations.
\newblock {\em Nature Communications}, 11:2385, 2020.

\bibitem{witte83}
C.~L. Witte and M.~H. Witte.
\newblock Circulatory dynamics of the spleen.
\newblock {\em Lymphology}, 16:60--71, 1983.

\bibitem{zebda13}
A.~Zebda et~al.
\newblock Single glucose biofuel cells implanted in rats power electronic
  devices.
\newblock {\em Scientific Reports}, 3:1516, 2013.

\end{thebibliography}
\end{document}